\documentclass[11pt]{article}

\usepackage[final]{acl}

\usepackage{times}
\usepackage{latexsym}

\usepackage[T1]{fontenc}

\usepackage[utf8]{inputenc}

\usepackage{microtype}

\usepackage{inconsolata}

\usepackage{graphicx}

\usepackage{amsmath}
\usepackage{amssymb}
\usepackage{bbding}
\usepackage{booktabs}
\usepackage{colortbl}
\usepackage{dsfont}
\usepackage{fontawesome5}
\usepackage{pifont}
\usepackage{makecell}
\usepackage{multirow}
\usepackage{xcolor}
\usepackage{enumitem}
\usepackage[most]{tcolorbox}

\newcommand\eat[1]{}

\newcommand{\ours}{\texttt{ReMe}}
\definecolor{lightblue}{RGB}{102, 167, 208}

\title{\textit{Remember Me, Refine Me}: A Dynamic Procedural Memory \\Framework for Experience-Driven Agent Evolution}

\author{Zouying Cao$^{1,3,}$\thanks{This work was done during Zouying Cao’s internship at Tongyi Lab,
Alibaba Group.}, Jiaji Deng$^{2}$, Li Yu$^{2}$, Weikang Zhou$^{2}$, \\
\textbf{Zhaoyang Liu}$^{2,}$\footnotemark[2], \textbf{Bolin Ding}$^{2}$, \textbf{Hai Zhao}$^{1,3,}$\thanks{Corresponding authors. This research was supported by the Shanghai Jiao Tong University 2030 Initiative and The Major Program of Chinese National Foundation of Social Sciences under Grant ‘The Challenge and Governance of Smart Media on News Authenticity’ [No. 23\&ZD213].} \\
\textsuperscript{1}AGI Institute, School of Computer Science, Shanghai Jiao Tong University, \\
\textsuperscript{2}Tongyi Lab, Alibaba Group, \\
\textsuperscript{3}Key Laboratory of Shanghai Education Commission for Intelligent Interaction\\ and Cognitive Engineering, Shanghai Jiao Tong University \\
\normalsize{\texttt{zouyingcao@sjtu.edu.cn, \{dengjiaji.djj,  jinli.yl, zhouweikang.zwk,}} \\
\normalsize{\texttt{jingmu.lzy, bolin.ding\}@alibaba-inc.com, zhaohai@cs.sjtu.edu.cn}} \\
}

\begin{document}
\maketitle

\begin{abstract}
Procedural memory enables large language model (LLM) agents to internalize ``how-to'' knowledge and thus reduce redundant trial-and-error. 
However, existing frameworks predominantly suffer from a ``passive accumulation'' paradigm, treating memory as a static append-only archive. 
To bridge the gap between static storage and dynamic reasoning, we propose \textbf{\ours} (\textit{Remember Me, Refine Me}), a comprehensive framework for experience-driven agent evolution. 
{\ours} manages the memory lifecycle via three mechanisms: 
1) \textit{multi-faceted distillation}, which extracts fine-grained experiences by recognizing success patterns, analyzing failure triggers and generating comparative insights; 
2) \textit{context-adaptive reuse}, which tailors historical insights to new contexts through scenario-aware indexing; 
and 3) \textit{utility-based refinement}, which automatically adds validated memories and prunes outdated ones to maintain a compact, high-quality experience pool. 
Experiments on BFCL-V3 and AppWorld demonstrate that {\ours} establishes a new state-of-the-art in agent memory system. 
Crucially, we observe a significant memory-scaling effect: Qwen3-8B equipped with {\ours} outperforms larger, memoryless Qwen3-14B, indicating that self-evolving memory provides a computation-efficient path for lifelong learning.\footnote{\url{https://github.com/agentscope-ai/ReMe}}
\end{abstract}

\section{Introduction}
\begin{figure*}[tbp]
    \centering
    \includegraphics[width=1\linewidth]{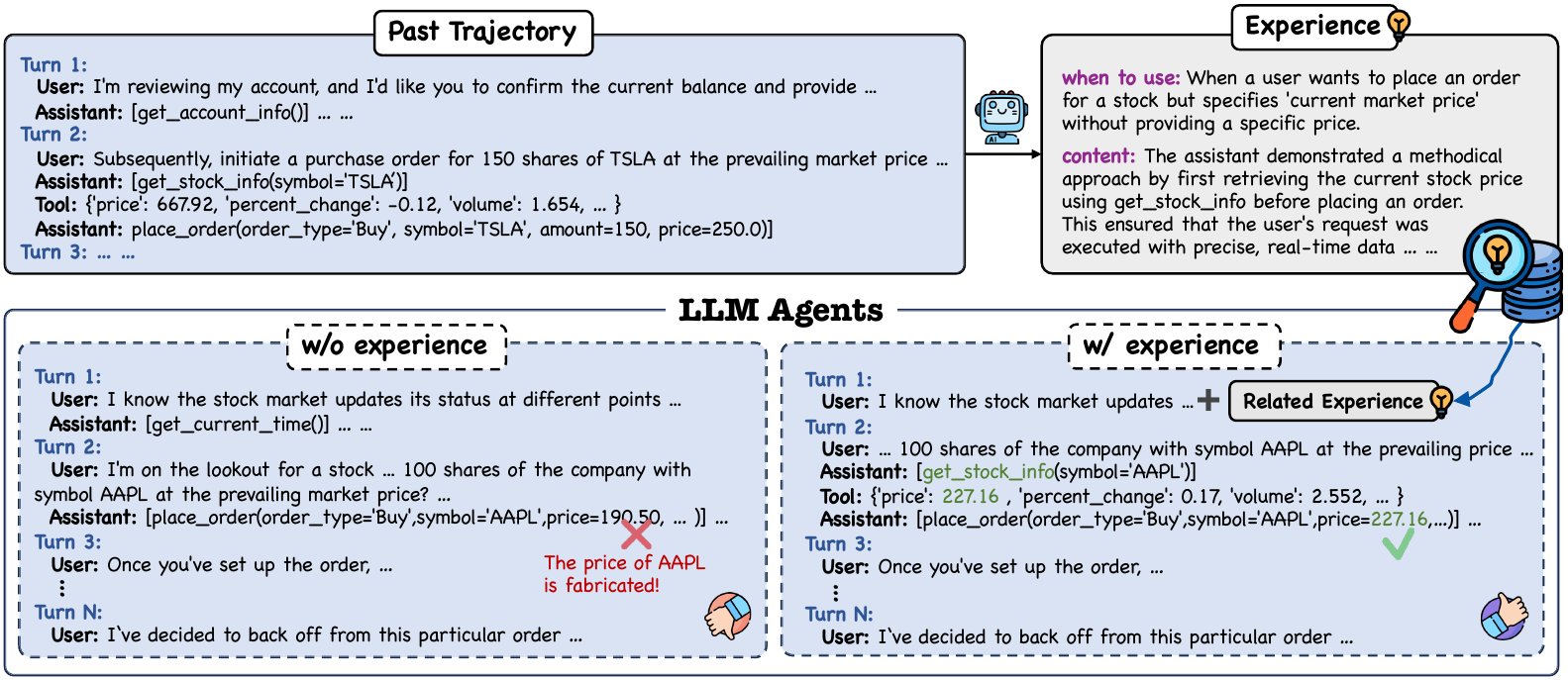}
    \caption{Example of how agents complete one stock trading task with and without past experience.}\label{fig:example}
\end{figure*}

The transition from static language models to autonomous agents marks a pivotal shift in artificial intelligence, enabling systems to handle complex, dynamic tasks through iterative reasoning and tool use~\citep{tao2024survey,gao2025survey,fang2025comprehensive}. 
To facilitate continuous improvement without model retraining, \textbf{procedural memory}, which internalizes ``how-to'' knowledge from past interactions, has emerged as a critical substrate for agent evolution~\citep{zhang2025survey,xu2025mem}. 
By accumulating high-quality problem-solving experiences, agents can leverage prior successes and lessons to navigate novel scenarios, theoretically reducing redundant trial-and-error and avoiding local optima~\citep{wang2025mirix,chen2025swe}. 
Figure~\ref{fig:example} contrasts how an agent completes one stock trading task with and without experiences.

To bridge the gap between static storage and dynamic reasoning, an ideal procedural memory system must function not merely as a database, but as an evolving cognitive substrate satisfying three core criteria: 
1) \textit{High-quality Extraction}: The system should distill generalized, reusable knowledge from noisy execution trajectories, rather than raw, problem-specific observations. 
2) \textit{Task-grounded Utilization}: Retrieved memories should be dynamically adapted to the specific requirements of the current task, maximizing their utility in novel scenarios.
3) \textit{Progressive Optimization}: The memory pool should maintain its vitality through continuous updates, autonomously reinforcing effective entries while removing outdated ones to prevent degradation over time.

However, current frameworks often fall short of these criteria, largely constrained by a ``passive accumulation'' paradigm. 
Prevailing approaches typically treat memory as inert, static storage, built on either raw trajectories as experiences~\citep{zheng2023synapse,hu2024hiagent} or summarized workflows corresponding to entire trajectories~\citep{tang2025agentkb,liu-etal-2025-contextual}. 
This introduces several fundamental limitations. 
First, coarse-grained trajectory-level experiences may introduce irrelevant information that can prevent the agent from grasping the core logic. 
Second, fetched experiences are applied without adaptation, leading to failures in slightly shifted scenarios. 
Crucially, lack of timely update strategies causes the experience pool to degrade into a mixture of valid insights and toxic noise~\citep{xiong2025memory}.

To address these challenges, we propose \textbf{\ours} (\textit{Remember Me, Refine Me}), a dynamic procedural memory framework that shifts the paradigm from passive storage to feedback-driven evolution. 
We introduce coordinated innovations across the memory lifecycle to meet the criteria of an ideal system.
First, {\ours} employs a multi-faceted distillation strategy for high-quality extraction. Through success pattern recognition, failure analysis and comparative insight generation, the system distills key steps from past execution trajectories into structured, reusable experiences. 
Second, we design a comprehensive reuse pipeline for task-grounded utilization. {\ours} employs usage scenario indexing strategy for retrieval, supplemented by reranking and adaptive rewriting, aligning historical insights with the specific constraints of new tasks. 
Finally, {\ours} implements a utility-based refinement mechanism for progressive optimization. 
The memory pool grows as new successful trajectories contribute reliable experiences and failure attempts trigger self-reflection to explore viable solutions for potential insights. 
Concurrently, our framework tracks the utility of each experience during reuse, periodically pruning low-utility entries to maintain a compact and highly effective memory state.

Through extensive experiments on BFCL-V3 and AppWorld benchmarks, {\ours} achieves state-of-the-art performance, demonstrating its effectiveness for memory-augmented agents. 
Most notably, results reveal that memory quality can substitute for model scale: {\ours} enables Qwen3-8B to outperform larger Qwen3-14B (without memory), achieving average gains of 8.83\% in Avg@4 and 7.29\% in Pass@4. 
These findings suggest that a self-evolving memory mechanism paves the way for resource-efficient lifelong learning in LLM agents.

In summary, our contributions are as follows:
\begin{itemize}[leftmargin=12pt,topsep=2pt,itemsep=2pt]
    \item We propose \textbf{\ours}, a comprehensive framework for agent evolution that integrates multi-faceted experience distillation, context-adaptive reuse, and utility-based refinement. This achieves the closed loop of procedural memory, resolving the ``passive accumulation'' dilemma by enabling agents to autonomously distill, adapt, and maintain high-quality reasoning patterns.
    \item We release \textbf{\texttt{reme.library}}, a fine-grained procedural memory dataset constructed from diverse agentic tasks, with structured success patterns and failure lessons to serve as a valuable community resource for studying procedure memory and optimizing memory-augmented agents.
    \item Extensive experiments show that {\ours} significantly enhances agent performance across diverse benchmarks. Crucially, we demonstrate a \textbf{memory-scaling effect}, where smaller models equipped with {\ours} surpass larger baselines, validating our framework as a computationally efficient pathway for lifelong agent learning.
\end{itemize}

\section{Related Works}
\textbf{Memory-enhanced LLM Agents.} 
LLM-based agents excel at handling complex tasks and interactions, fueling their integration into diverse fields, such as finance~\citep{ding2024large}, education~\citep{wang2024large} and personalized assistant applications~\citep{abbasian2023conversational}.
Contemporary LLM agents employ memory systems that store explored information and reuse these experiences, to enhance their reasoning capabilities and training efficiency~\citep{mei2025survey}.
In general, memory-enhanced agents often leverage two forms of memory: parametric memory and non-parametric memory~\citep{zhang2024survey}. 
Parametric memory refers to encoding long-term knowledge within model weights, while non-parametric memory utilizes external resources like knowledge bases and databases to enrich task contexts without modifying model parameters.
WKM~\citep{qiao2024agent} incorporates a parametric world-knowledge model to facilitate agent planning. 
AWM~\citep{wang2025agent} enables agents to automatically induce and use task workflows from past experiences, improving their performance on web navigation tasks.
MARK~\citep{ganguli2025mark} constructs user preference memory to deliver personalized responses in conversational AI systems.

\noindent
\textbf{Experience Learning Strategies.}
Recent studies show LLMs can improve their decision-making abilities through gathering experiences and recalling relevant knowledge~\citep{zhao2024expel,tan-etal-2025-prospect}. 
The core of experience learning involves extracting usable information to selectively update the experience pool and retrieving effective experiences to help generate responses. 
Early approaches, such as Synapse~\citep{zheng2023synapse} and HiAgent~\citep{hu2024hiagent}, store complete trajectories as experiences for retrieval. 
However, collecting raw and long interaction histories is hard to manage, and the lack of abstraction limits task generalization. 
Current works~\citep{wang2025agent,chen2025swe} focus on summarizing structured knowledge from prior trajectories and implementing a context-aware retrieval system to reuse experiences for task guidance. 
For instance, Agent KB~\citep{tang2025agentkb} captures generalizable experience units and introduces a teacher-student dual-phase retrieval mechanism that enables complex agentic problem solving. 
CER~\citep{liu-etal-2025-contextual} distills fine-grained skills and environment dynamics, allowing agents to augment themselves with relevant knowledge in new tasks. 
These methods 
neglect strategic experience removal mechanism, since harmful experiences inevitably exist even with human validation and initial helpful ones can also degrade over time~\citep{xiong2025memory}. 

\begin{figure*}[tbp]
    \centering
    \includegraphics[width=1\linewidth]{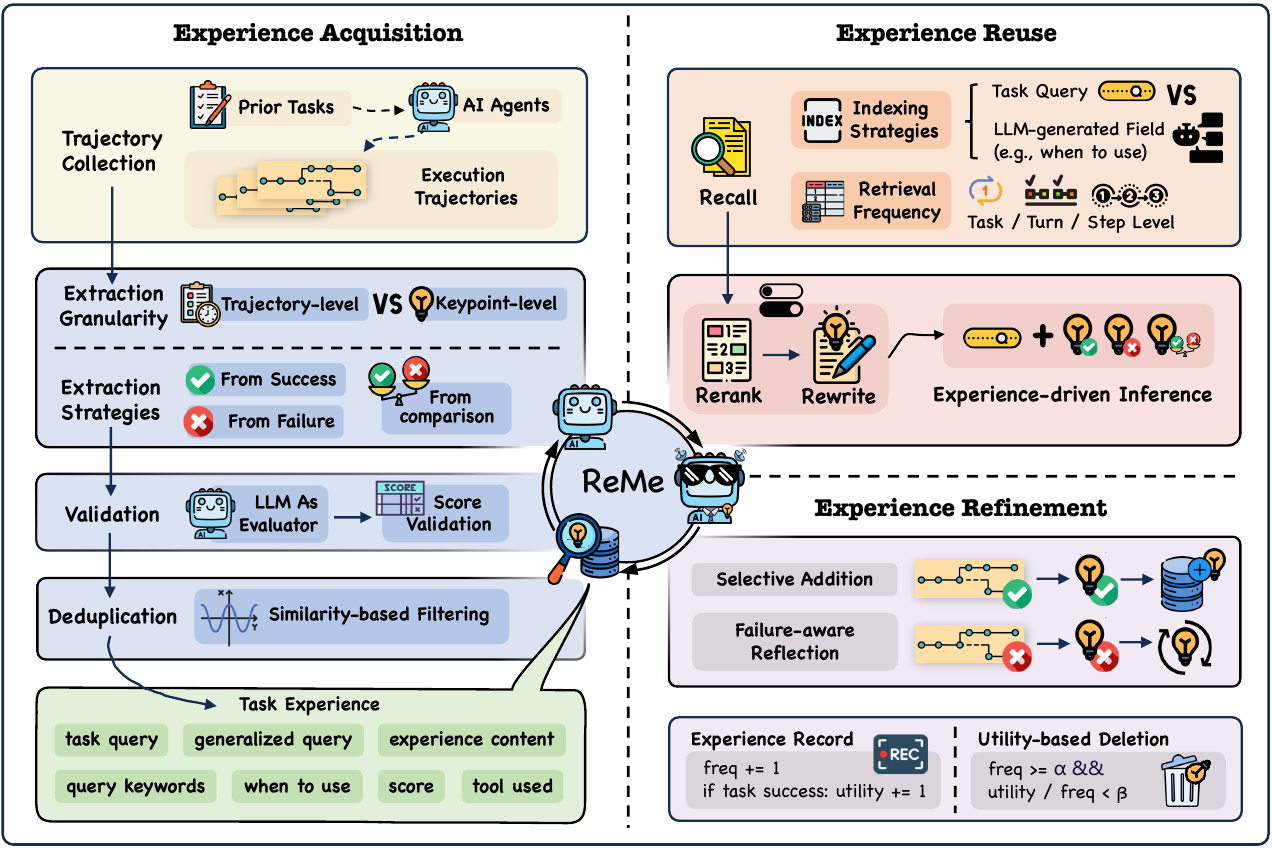}
    \caption{The {\ours} framework comprises three alternating phases: build an experience pool from the agent's past trajectories, recall and adapt relevant experiences for new tasks, then refine the pool by selectively adding new insights and removing outdated ones after each task execution.}\label{fig:overview}
\end{figure*}
\section{Methodology}\label{method}
\subsection{Overview of \ours}
Our framework, \ours, as illustrated in Figure~\ref{fig:overview}, operates through three interconnected phases: experience acquisition, reuse, and refinement. 
In the \textbf{\textit{experience acquisition}} phase, a summarizer analyzes agent generated trajectories (both successful and failed) and distills actionable knowledge into a structured experience pool. 
During \textbf{\textit{experience reuse}}, given a novel task, a retriever recalls relevant experiences from the experience pool. These experiences then augment the agent's context, enhancing their reasoning and task-solving performance. 
Finally, the \textbf{\textit{experience refinement}} phase continuously optimizes the experience pool by incorporating new solid experiences and discarding outdated ones, ensuring long-term relevance and adaptability to shifting task demands.

\subsection{Experience Acquisition}
We begin by defining agentic experiences $\mathcal{E}$ as structured, generalizable representations of agent execution insights. 
Each individual experience $E\in\mathcal{E}$ is denoted as $E=\langle \omega, e, \kappa,c,\tau \rangle$, where $\omega$ states the scenario when to use the experience, $e$ represents the core experience content, $\kappa = \{\kappa_1, \kappa_2, ..., \kappa_m\}$ is a set of relevant keywords for categorization, $c \in [0,1]$ quantifies the confidence score, 
and $\tau$ enumerates the tools utilized.

To construct the initial experience pool, the execution agent $\text{LLM}_{execute}$ interacts with the environment over time and across the training tasks, incrementally accumulating informative trajectories. 
For each task query $q$, we sample trajectories $N$ times aiming to capture diverse execution paths and thereby increase the likelihood of obtaining valuable success/failure pairs for comparisons during \textbf{\textit{experience acquisition}}.

After collecting a set of exploration trajectories, a summarizer $\text{LLM}_{summ}$ is instructed to transform them into structured, reusable experiences through three complementary analyses: 
First, the summarizer engages in success pattern recognition, identifying effective strategies and distilling the underlying principles from succeeded trajectories. 
Concurrently, $\text{LLM}_{summ}$ conducts failure analysis, scrutinizing unsuccessful attempts to derive valuable lessons. 
These preventive insights discuss common pitfalls, ineffective approaches, and critical errors that can be used to avoid repeating them in future tasks.  
Additionally, $\text{LLM}_{summ}$ performs comparative analysis by jointly examining successful and failed trajectories, identifying critical differences that distinguish effective from ineffective attempts.

Following the summarization, a validation step leveraging LLM-as-a-Judge~\citep{zheng2023judging} is further applied to assess whether the extracted experiences are actionable, accurate, and valuable for future agent executions. 
The designed prompt template is presented in Appendix Table~\ref{tab:validation_prompt}. 
Moreover, to keep the experience pool compact, validated experiences undergo similarity‑based deduplication. 
Specifically, each new experience is encoded into a vector embedding and compared against existing ones via cosine similarity, and any candidate exceeding a predefined similarity threshold is discarded as redundant.
This helps maintain the efficiency of the subsequent experience reuse phase and preserve the diversity of retrieved experiences.

All retained experiences are indexed by the embedding vector of usage scenario $\omega$ and then stored in a vector database, which we refer to as the experience pool. 
The multi-faceted experience pool establishes a foundation for efficient retrieval and application of relevant knowledge in future problem-solving scenarios, promoting the agent evolution from trial-and-error to strategic reasoning.

\subsection{Experience Reuse}\label{sec:reuse}

Equipped with the experience pool, we can retrieve top-$K$ relevant experiences based on task similarity, which serve as a candidate set of in-context learning demonstrations to guide $\text{LLM}_{execute}$. 
To be specific, the retriever utilizes advanced embedding models (e.g., Qwen3-Embedding~\citep{qwen3embedding}) to encode the current task query and computes cosine similarity scores to rank prior experiences. 
More retrieval details can be found in Appendix~\ref{app:retrieval}. 
Upon fetching the top-$K$ experiences, we optionally employ a context-aware reranker $\text{LLM}_{rerank}$ to further refine the selection. 
This involves a nuanced evaluation of experience relevance in light of the current task's specific context, constraints, and objectives, thus ensuring the most pertinent experiences are brought to the forefront. 
Table~\ref{tab:rerank_prompt} shows the prompt template.

To better adapt the experiences to new task requirements, we introduce the rewriting module to reorganize the original context (containing multiple experiences) into a cohesive, task-specific guidance that is more directly applicable. 
See Table~\ref{tab:rewrite_prompt} for the example prompt. 
Since past experiences may not always perfectly align with new situations, this intelligent adaptation mechanism not only increases the immediate utility of the retrieved experiences but also empowers the agent to make more flexible and context-aware decisions.


The \textbf{\textit{experience reuse}} phase extends beyond mere experience retrieval, acting as a cognitive bridge that dynamically connects past knowledge with present challenges. 
By combining retrieval, reranking and rewriting, it not only leverages prior wisdom but also encourages novel thinking when past experiences fall short, thereby achieving a balance between exploitation and exploration.

\subsection{Experience Refinement}\label{sec:refinement}
However, a static experience pool cannot adapt to shifts in task distributions or improvements in model capability, making retrieved experiences increasingly irrelevant.
To address this, we introduce an \textbf{\textit{experience refinement}} mechanism that dynamically updates the experience pool via selective addition and utility‑based deletion. 

First, we carefully compare two distinct strategies for adding new experiences to the pool: 
1) full addition, which incorporates experiences summarized from all new trajectories regardless of outcome; 2) selective addition, where only trajectories that lead to success are distilled into experiences and stored. 
The empirical evidence indicates that full addition often underperforms selective addition, which may be attributed to the quality of failure-based experiences. 
During initial experience pool construction, multiple failed trajectories can be collectively analyzed to extract meaningful insights. 
However, in real-time task execution, a single failed trajectory often provides insufficient context for accurate failure analysis, potentially leading to misguided experiences. 
In contrast, successful trajectories consistently yield more reliable and actionable insights, thereby making selective addition effective.

Additionally, we recognize the potential value of learning from failures and introduce a failure-aware reflection mechanism that encourages agents to explore alternative strategies when encountering new task failures. 
Specifically, $\text{LLM}_{summ}$ analyzes this unsuccessful attempt, extracts key insights about potential areas for improvement, and then $\text{LLM}_{execute}$ starts a new trial based on these lessons. 
When such trial succeeds, the corresponding lessons are incorporated into memory; otherwise, they are discarded without cluttering the experience pool. 
To avoid falling into an endless loop caused by inherent model limitations, we limit the maximum number of self-reflections to 3. 

Second, to prevent the accumulation of outdated or ineffective experiences, we employ a utility-based deletion strategy that removes any experience whose average utility across all its past recalls falls below a predefined threshold $\beta$. 
Specifically, {\ours} continuously records the status of existing experiences, including the total retrievals $f$ and the historical utility $u$ which increments by 1 each time its recall contributes to a successful task completion. 
An experience $E\in \mathcal{E}$ is considered to be removed when it is frequently retrieved yet fails to improve new task performance: 
\begin{equation}
\phi_{remove}(E) = 
\begin{cases}
\mathds{1} \left[\frac{u(E)}{f(E)}  \leq \beta\right], & \text{if }f(E)\geq \alpha, \\
0\;, & \text{otherwise}.
\end{cases}
\end{equation}
Note that we only consider an experience for removal after it has been retrieved at least $\alpha$ times. 

By integrating these components, {\ours} facilitates a self-evolving experience pool that retains high-quality experiences capable of shaping long-term agent behavior while adapting to new task demands.

\begin{table*}
    \centering
    \resizebox{\linewidth}{!}{
    \begin{tabular}{llcccccc}
    \toprule
         \multirow{2}{*}{\textbf{Model}} & \multirow{2}{*}{\textbf{Methods}} & \multicolumn{2}{c}{\textbf{BFCL-V3}} & \multicolumn{2}{c}{\textbf{AppWorld}} & \multicolumn{2}{c}{\textbf{Avg}} \\ 
         &  & Avg@4 & Pass@4& Avg@4 & Pass@4& Avg@4 & Pass@4  \\ 
    \midrule
        \multirow{5}{*}{\textbf{Qwen3-8B}} & No Memory & 40.33$^{\textcolor{lightblue}{\pm0.94}}$ & 59.55$^{\textcolor{lightblue}{\pm0.83}}$ & 14.97$^{\textcolor{lightblue}{\pm0.24}}$& 32.85$^{\textcolor{lightblue}{\pm2.11}}$&
        27.65& 46.20\\ 
         & A-Mem & 41.22$^{\textcolor{lightblue}{\pm0.61}}$ & 62.00$^{\textcolor{lightblue}{\pm2.37}}$& 12.95$^{\textcolor{lightblue}{\pm0.37}}$ & 29.76$^{\textcolor{lightblue}{\pm2.80}}$ &27.09&45.88 \\ 
         & LangMem & 44.11$^{\textcolor{lightblue}{\pm0.28}}$ & 65.55$^{\textcolor{lightblue}{\pm1.13}}$ & 11.46$^{\textcolor{lightblue}{\pm0.53}}$ & 26.79$^{\textcolor{lightblue}{\pm0.84}}$ & 27.79 & 46.17 \\ 
         & {\ours} \texttt{(fixed)} & 44.50$^{\textcolor{lightblue}{\pm0.85}}$& 65.77$^{\textcolor{lightblue}{\pm0.63}}$ & 17.06$^{\textcolor{lightblue}{\pm0.25}}$& 36.31$^{\textcolor{lightblue}{\pm1.29}}$ & 30.78&51.04\\ 
         & \cellcolor{gray!15}{\ours} \texttt{(dynamic)}& \cellcolor{gray!15}\textbf{45.17}$^{\textcolor{lightblue}{\pm0.36}}$& \cellcolor{gray!15}\textbf{68.00}$^{\textcolor{lightblue}{\pm0.55}}$  & \cellcolor{gray!15}\textbf{24.70}$^{\textcolor{lightblue}{\pm1.04}}$ & \cellcolor{gray!15}\textbf{42.06}$^{\textcolor{lightblue}{\pm0.74}}$  & \cellcolor{gray!15}\textbf{34.94} & \cellcolor{gray!15}\textbf{55.03} \\ 
    \midrule
        \multirow{5}{*}{\textbf{Qwen3-14B}} & No Memory  & 48.66$^{\textcolor{lightblue}{\pm1.51}}$& 68.22$^{\textcolor{lightblue}{\pm0.63}}$ & 22.57$^{\textcolor{lightblue}{\pm0.19}}$& 41.07$^{\textcolor{lightblue}{\pm0.84}}$ & 35.62&54.65\\ 
         & A-Mem & 47.44$^{\textcolor{lightblue}{\pm0.44}}$& 69.77$^{\textcolor{lightblue}{\pm0.63}}$ & 18.95$^{\textcolor{lightblue}{\pm0.31}}$ & 37.70$^{\textcolor{lightblue}{\pm0.57}}$ & 33.20&53.74\\ 
         & LangMem & 49.17$^{\textcolor{lightblue}{\pm0.33}}$& 71.33$^{\textcolor{lightblue}{\pm1.33}}$ & 21.88$^{\textcolor{lightblue}{\pm1.37}}$& 41.67$^{\textcolor{lightblue}{\pm1.68}}$ & 35.53&56.50 \\ 
         & {\ours} \texttt{(fixed)} & 51.89$^{\textcolor{lightblue}{\pm0.34}}$& 72.44$^{\textcolor{lightblue}{\pm0.63}}$ & 25.35$^{\textcolor{lightblue}{\pm0.91}}$& 46.82$^{\textcolor{lightblue}{\pm0.74}}$ & 38.62&59.63  \\ 
         & \cellcolor{gray!15}{\ours} \texttt{(dynamic)} & \cellcolor{gray!15}\textbf{55.00}$^{\textcolor{lightblue}{\pm0.72}}$ & \cellcolor{gray!15}\textbf{74.44}$^{\textcolor{lightblue}{\pm0.83}}$  & \cellcolor{gray!15}\textbf{34.32}$^{\textcolor{lightblue}{\pm0.81}}$  & \cellcolor{gray!15}\textbf{52.98}$^{\textcolor{lightblue}{\pm1.29}}$ &\cellcolor{gray!15}\textbf{44.66}&\cellcolor{gray!15}\textbf{63.71} \\ 
    \midrule
        \multirow{5}{*}{\textbf{Qwen3-32B}} & No Memory & 54.55$^{\textcolor{lightblue}{\pm0.63}}$ & 72.44$^{\textcolor{lightblue}{\pm0.83}}$ & 27.23$^{\textcolor{lightblue}{\pm0.92}}$ & 50.59$^{\textcolor{lightblue}{\pm1.68}}$ & 40.89&61.52 \\ 
         & A-Mem & 54.50$^{\textcolor{lightblue}{\pm1.09}}$& 72.66$^{\textcolor{lightblue}{\pm0.54}}$ &28.13$^{\textcolor{lightblue}{\pm0.75}}$& 51.19$^{\textcolor{lightblue}{\pm0.97}}$ & 41.32&61.93\\ 
         & LangMem & 52.27$^{\textcolor{lightblue}{\pm1.13}}$& 72.22$^{\textcolor{lightblue}{\pm1.91}}$& 24.55$^{\textcolor{lightblue}{\pm0.57}}$& 47.02$^{\textcolor{lightblue}{\pm1.56}}$& 38.41 & 59.62 \\ 
         & {\ours} \texttt{(fixed)} & 56.05$^{\textcolor{lightblue}{\pm1.26}}$& 74.89$^{\textcolor{lightblue}{\pm0.63}}$  & 31.50$^{\textcolor{lightblue}{\pm0.67}}$& 58.13$^{\textcolor{lightblue}{\pm1.40}}$&43.78&66.51 \\ 
         & \cellcolor{gray!15}{\ours} \texttt{(dynamic)} & \cellcolor{gray!15}\textbf{56.17}$^{\textcolor{lightblue}{\pm0.24}}$& \cellcolor{gray!15}\textbf{76.44}$^{\textcolor{lightblue}{\pm1.13}}$ & \cellcolor{gray!15}\textbf{42.02}$^{\textcolor{lightblue}{\pm0.51}}$& \cellcolor{gray!15}\textbf{63.49}$^{\textcolor{lightblue}{\pm0.28}}$ & \cellcolor{gray!15}\textbf{49.10} & \cellcolor{gray!15}\textbf{69.97} \\ 
    \bottomrule
    \end{tabular}
    }
    \caption{Performance comparison (\%) between {\ours} and the baselines on BFCL-V3, AppWorld benchmarks. \textbf{Bold} indicate the best results of each model. All results are computed as the average over three independent runs, with the superscript showing the standard deviation.}\label{tab:main}
\end{table*}

\section{Experiments}
\subsection{Experimental Settings}
\paragraph{Datasets.} 
We conduct experiments on two tool-augmented benchmarks: 
BFCL-V3~\citep{patilberkeley}, AppWorld~\citep{trivedi2024appworld}. 
For BFCL-V3, we randomly select 50 tasks from the base multi-turn category to construct the initial experience pool since the default dataset does not provide training split.  
The remaining 150 tasks serve as the evaluation set. 
For AppWorld, we use 90 training tasks for the initial experience acquisition stage and evaluate agents on 168 test-normal tasks. 
Detailed information of the datasets are in Appendix~\ref{app:dataset}. 

\paragraph{Metrics.} We report both Avg@4 and Pass@4 results: the average task success rate across four independent trials, and the probability that at least one out of four independent task trials is successful. 
Unless otherwise specified, all results are averaged over three independent runs and reported as mean with standard deviation. 

\paragraph{Baselines.} 
To evaluate the effectiveness of {\ours}, we compare it against three baselines: (1) No Memory, and two popular baseline memory systems (2) A-Mem~\citep{xu2025mem}, an agentic memory system that enables LLM agents to dynamically organize their memories for future action guidance, and (3) LangMem~\citep{LangChain2025}, LangChain’s long-term memory module that provides tooling to extract important information from previous conversations and optimize agent behavior through prompt refinement. 
For fair comparison, all methods perform experience retrieval only once at the beginning of each task. Additionally, the memory addition operation for these systems is triggered only upon the collection of successful trajectories.
Further implementation details of the baseline methods are provided in Appendix~\ref{app:baselines}.

\paragraph{Implementation Details.} 
We use the Qwen3 series instruct models~\citep{qwen3technicalreport} as $\text{LLM}_{execute}$ and set $\text{LLM}_{summ}=\text{LLM}_{execute}$ for experience-driven self-evolution. 
In the experience acquisition phase, we set $N$=8 and temperature=0.9 for trajectory sampling. 
For experience indexing, we employ \textit{text-embedding-v4}~\footnote{https://bailian.console.aliyun.com/?tab=model\#/model-market/detail/text-embedding-v4} with its default embedding dimension of 1024. 
The prompts used in this phase and more details can be found in Appendix~\ref{app:settings}. 
In the experience reuse phase, we use top-$K$=5, retrieving the five most relevant experiences for each task. 
The configuration difference between {\ours} \texttt{(fixed)} and {\ours} \texttt{(dynamic)} lies in whether the experience pool is dynamically updated during agent execution. 
In the experience refinement phase, utility-based deletion is controlled by the retrieval threshold $\alpha$=5 and the utility threshold $\beta$=0.5, where the threshold selection follows the prior work~\citep{xiong2025memory}. 
Selecting $3$ for the maximum number of self-reflections is also proper since the agent's performance immediately improves between the first two trials~\citep{shinn2023reflexion}. 
Additionally, the maximum number of iterations is limited to 30, after which the agent terminates regardless of task success or failure.
To ensure fair comparison, we maintain these settings consistently across all experiments unless otherwise specified for ablation studies. 

\subsection{Main Results} 
Table~\ref{tab:main} presents the main results of {\ours} across Qwen3 family models on BFCL-V3 and AppWorld benchmarks. 
Overall, {\ours} achieves the highest average task success rate across three model sizes, consistently outperforming No Memory baseline and competitive baseline memory systems. 
Specifically, Qwen3-8B with {\ours} surpasses the No Memory baseline by an improvement of 7.29\% Pass@4 and 8.83\% Avg@4 on average. 
The gains observed in Pass@4 indicate that retrieved experiences are effective at broadening the exploration space, increasing the likelihood of finding at least one successful solution among multiple attempts.
Besides, the performance stability of our {\ours} is particularly evident when compared to the baseline methods. 
For instance, while LangMem performs well on BFCL-V3, its performance drops significantly on AppWorld, especially for smaller models. 
Instead, {\ours} \texttt{(dynamic)} shows remarkable consistency across both BFCL-V3 and AppWorld benchmarks.

\begin{figure}
    \centering
    \includegraphics[width=1\linewidth]{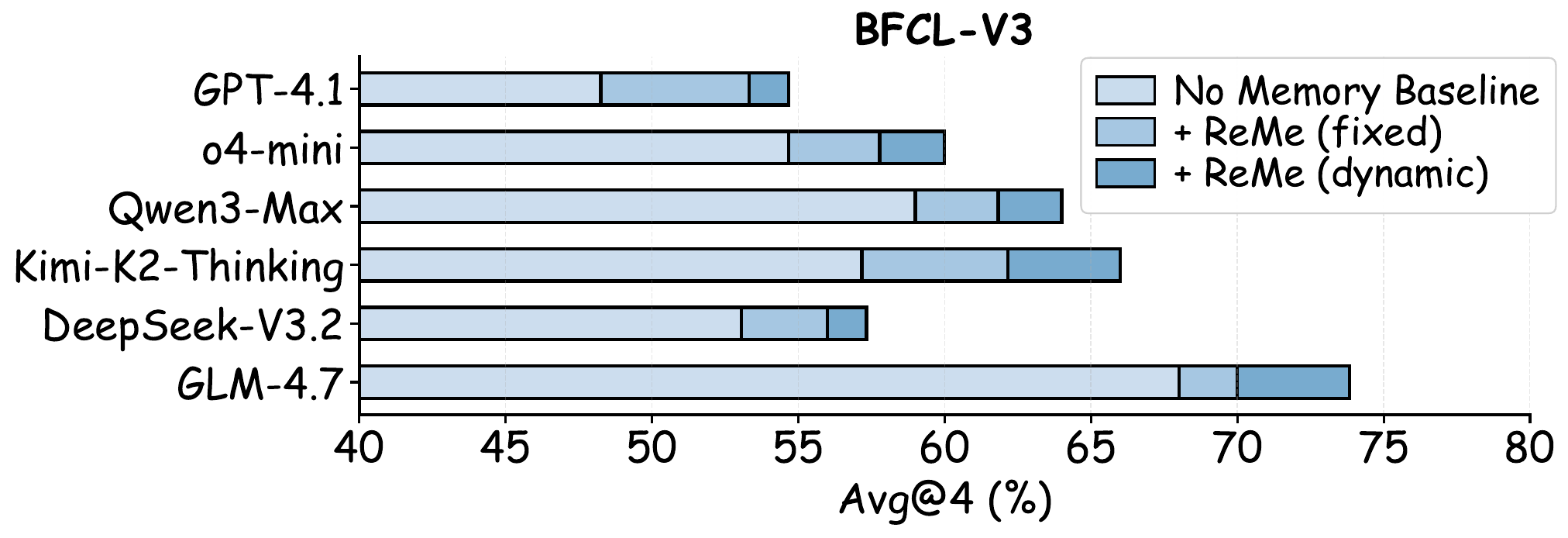}
    \caption{Performance improvements (\%) for multiple LLMs enhanced with {\ours}.}
    \label{fig:more_models}
\end{figure}

\begin{table}[tbp]
\centering
\setlength\tabcolsep{3pt}
\resizebox{\linewidth}{!}{
    \begin{tabular}{lcccc}
    \toprule
    \multirow{2}{*}{\textbf{\makecell[c]{Granularity}}} &\multicolumn{2}{c}{\textbf{Qwen3-8B}}&\multicolumn{2}{c}{\textbf{Qwen3-14B}}\\ 
    & Avg@4(\%) & Pass@4(\%)& Avg@4(\%)& Pass@4(\%)\\ 
    \midrule
    \textbf{Trajectory-level} &43.00$_{\textcolor{red}{+2.67}}$&60.00$_{\textcolor{red}{+0.45}}$& 49.66$_{\textcolor{red}{+1.00}}$&69.33$_{\textcolor{red}{+1.11}}$  \\ 
    \rowcolor{lightblue!20}\textbf{Keypoint-level} & 44.50$_{\textcolor{red}{+4.17}}$ & 65.77$_{\textcolor{red}{+6.22}}$&51.89$_{\textcolor{red}{+4.23}}$&72.44$_{\textcolor{red}{+4.22}}$\\
    \bottomrule
    \end{tabular}
}
\caption{Ablation study on extraction granularity levels in the experience acquisition stage. The experimental setting is {\ours} \texttt{(fixed)}, with subscript showing the performance gap compared with No Memory baseline.}\label{tab:keypoint}
\end{table}

Notably, smaller models equipped with {\ours} can be comparable to, or even surpass, larger models without memory. 
For example, the average Pass@4 score for Qwen3-8B + {\ours} \texttt{(dynamic)} exceeds that of vanilla Qwen3-14B (55.03\% vs. 54.65\%). 
Similarly, Qwen3-14B + {\ours} \texttt{(dynamic)} exceeds the overall performance of Qwen3‑32B without memory (Avg@4: 44.66\% vs. 40.89\%; Pass@4: 63.71\% vs. 61.52\%). 
This underscores that an effective memory mechanism can significantly narrow the performance gap across model scales. 

Moreover, {\ours} \texttt{(dynamic)} consistently outperforms {\ours} \texttt{(fixed)} across all model sizes and benchmarks. This underscores the importance of adaptive experience refinement during task execution. 
In addition, {\ours} tends to reduce the standard deviation in performance across runs, particularly for larger models. This suggests that {\ours} not only improves overall performance but also enhances the robustness and reliability of model outputs.

To demonstrate the generalizability of {\ours} across different backbones, we evaluate on six additional LLMs as shown in Figure~\ref{fig:more_models}. {\ours} delivers consistent gains in Avg@4 across all models, with the dynamic variant yielding further improvements. 

\subsection{Ablation Studies}\label{sec:ablation}
\paragraph{Granularity Ablations.} 
We compare two granularity levels for experience  acquisition: trajectory-level and keypoint-level. 
In Appendix~\ref{app:example}, we present two experience examples illustrating the structural and content differences between these granularity settings. 
As shown in Table~\ref{tab:keypoint}, although the incorporation of trajectory-level experiences exhibits minor progress over No Memory baseline, the performance gains brought by keypoint-level experiences are substantially higher. 
This underscores that summarizing experiences at a fine-grained level enables more effective knowledge transfer, leading to superior agent performance across different tasks and model scales. 

\begin{table}[tbp]
    \centering
    \setlength\tabcolsep{4pt}
    \resizebox{\linewidth}{!}{
    \begin{tabular}{@{}cccccc@{}}
    \toprule
    \multirow{2}{*}{\textbf{\makecell[c]{Full\\Addition}}} & \multirow{2}{*}{\textbf{\makecell[c]{Selective\\Addition}}} & \multirow{2}{*}{\textbf{\makecell[c]{Reflection}}} & \multirow{2}{*}{\textbf{\makecell[c]{Deletion}}} & \multicolumn{2}{c}{\textbf{BFCL-V3}}  \\
         &  & & & Avg@4 & Pass@4\\ 
    \midrule
        \checkmark & -- & -- & -- & 40.83\% & 62.00\% \\ 
         -- & \checkmark & -- & -- & 44.33\% & 64.66\%\\ 
         -- & \checkmark & \checkmark & -- & 45.00\% & 64.66\% \\ 
         -- & \checkmark & \checkmark & \checkmark & \textbf{45.17\%} & \textbf{68.00\%}\\ 
    \bottomrule
    \end{tabular}
    }
    \caption{Ablation on key components. We compare the full addition and selective addition and assess the impact of failure-aware reflection and utility-based deletion. A checkmark (\checkmark) indicates the component is used. }\label{tab:component_ablation}
\end{table}

\begin{table}[tbp]
    \centering
    \resizebox{\linewidth}{!}{
    \begin{tabular}{ccccc}
    \toprule
    \multirow{2}{*}{\textbf{\makecell[c]{Qwen3-8B}}} & \multirow{2}{*}{\textbf{\makecell[c]{Rerank}}} & \multirow{2}{*}{\textbf{\makecell[c]{Rrewrite}}} & \multicolumn{2}{c}{\textbf{BFCL-V3}}  \\
         & & & Avg@4 & Pass@4\\ 
    \midrule
    No Memory & -- & -- & 24.41\% & 28.50\% \\ 
    \midrule[0.05pt]
    \multirow{4}{*}{+{\ours}} &  -- & -- & 27.17\% & 34.66\%\\ 
     & \checkmark & -- & 28.91\% & 36.67\% \\ 
     & -- & \checkmark & 28.67\% & 37.33\%\\ 
     & \checkmark & \checkmark & \textbf{29.00\%} & \textbf{40.67\%}\\ 
    \bottomrule
    \end{tabular}
    }
    \caption{Ablation on the reranking and rewriting module. A checkmark (\checkmark) indicates the component is used. }\label{tab:component_ablation1}
\end{table}

\paragraph{Component Ablations.} 
Taking Qwen3-8B as an example, Table~\ref{tab:component_ablation} presents an ablation study on key components of our {\ours} framework. 
Firstly, replacing full addition with selective addition leads to substantial performance improvements, with gains of 3.50\% Avg@4 and 2.66\% Pass@4 on BFCL-V3. 
This highlights the importance of experience quality over quantity in experience-driven agent evolution. 
Moreover, the introduction of the failure-aware reflection module enhances the average task success rate, demonstrating the value of learning from unsuccessful attempts. 
Notably, incorporating the utility-based deletion yields further improvements, indicating that regularly discarding outdated experiences is critical for agents to adapt to non-stationary environments. 
Further, we supplement the ablation studies on the reranking and rewriting module. 
All tests use the Qwen3-8B model with thinking mode disabled in \ours \texttt{(fixed)} setting, with BFCL-V3 results summarized in Table~\ref{tab:component_ablation1}.

\begin{figure}[tbp]
    \centering
    \includegraphics[width=1\linewidth]{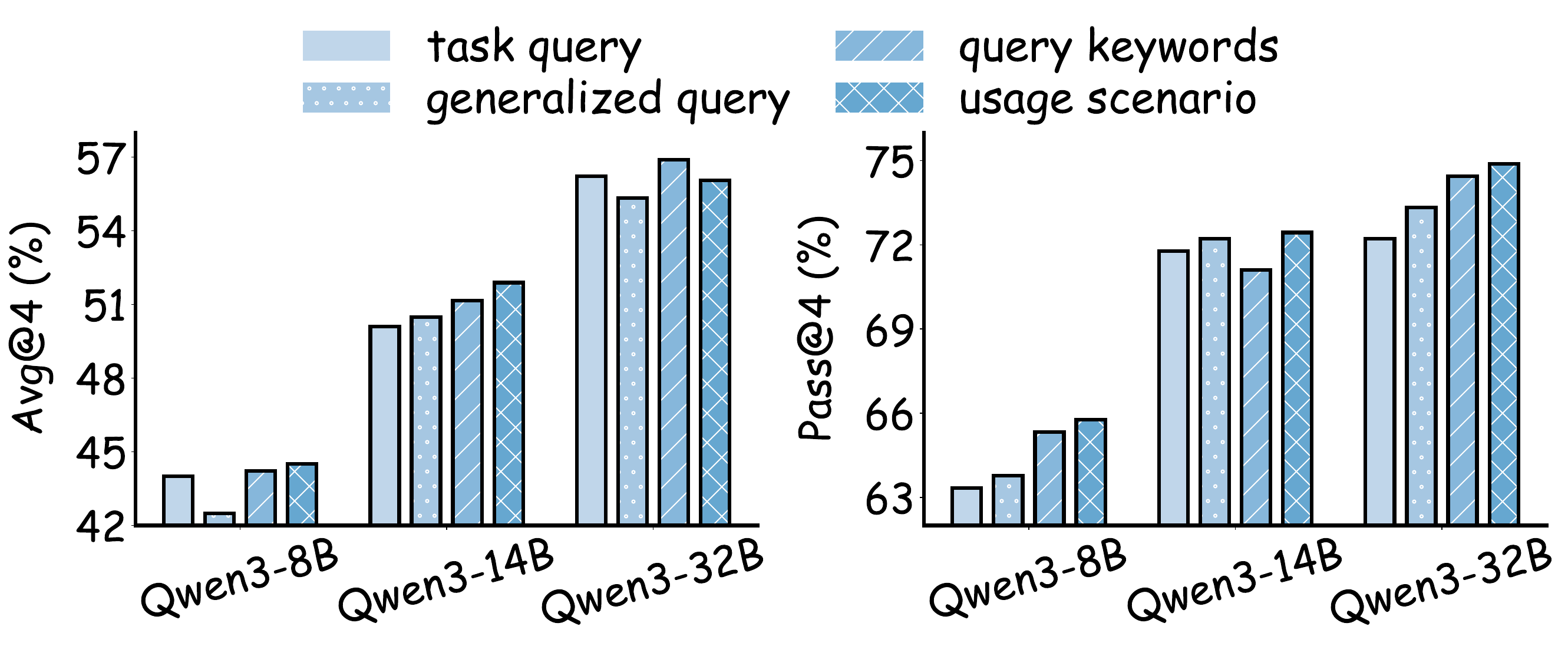}
    \caption{Ablation on retrieval keys. The experiments are evaluated on BFCL-V3 in {\ours} \texttt{(fixed)} setting.}
    \label{fig:key}
\end{figure}
\begin{table}[t]
\centering
\setlength\tabcolsep{4pt}
\resizebox{\linewidth}{!}{
    \begin{tabular}{l|l|cc}
    \toprule
    \multirow{2}{*}{\textbf{${\text{LLM}}_{\text{execute}}$}} &\multirow{2}{*}{\textbf{${\text{LLM}}_{\text{summ}}$}} &\multicolumn{2}{c}{\textbf{BFCL-V3}}\\ 
    && Avg@4 (\%)& Pass@4 (\%)\\ 
    \midrule
    \multirow{3}{*}{Qwen3-8B} & Qwen3-8B & 44.50 & 65.77 \\ 
    &Qwen3-14B & 46.33 \colorbox{lightblue!20}{$\textcolor{darkblue}{\triangle=1.83 \uparrow}$} & 66.00 \colorbox{lightblue!20}{$\textcolor{darkblue}{\triangle=0.23 \uparrow}$} \\
    &Qwen3-32B & 47.83 \colorbox{lightblue!20}{$\textcolor{darkblue}{\triangle=3.33 \uparrow}$} & 68.00 \colorbox{lightblue!20}{$\textcolor{darkblue}{\triangle=2.23 \uparrow}$}  \\
    \bottomrule
    \end{tabular}
}
\caption{Performance of different ${\text{LLM}}_{summ}$ capabilities with fixed ${\text{LLM}}_{execute}$ in {\ours} \texttt{(fixed)} setting. }\label{tab:summarizer}
\end{table}

\paragraph{Retrieval Key Ablations.} Regarding the indexing strategy, we explore four different retrieval keys to assess their impact on the performance of {\ours}. From Figure~\ref{fig:key}, it can be seen that using the raw task description or their extracted keywords to index experiences underperforms the LLM-generated fields (generalized query and usage scenario). 
The usage scenario indexing strategy, which likely captures both the task context and potential application areas, proves to be the most effective in retrieving relevant experiences from the database. 
For comprehensive results, please refer to Appendix~\ref{app:retrieval_key}.

\begin{figure}[t]
    \centering
    \includegraphics[width=1\linewidth]{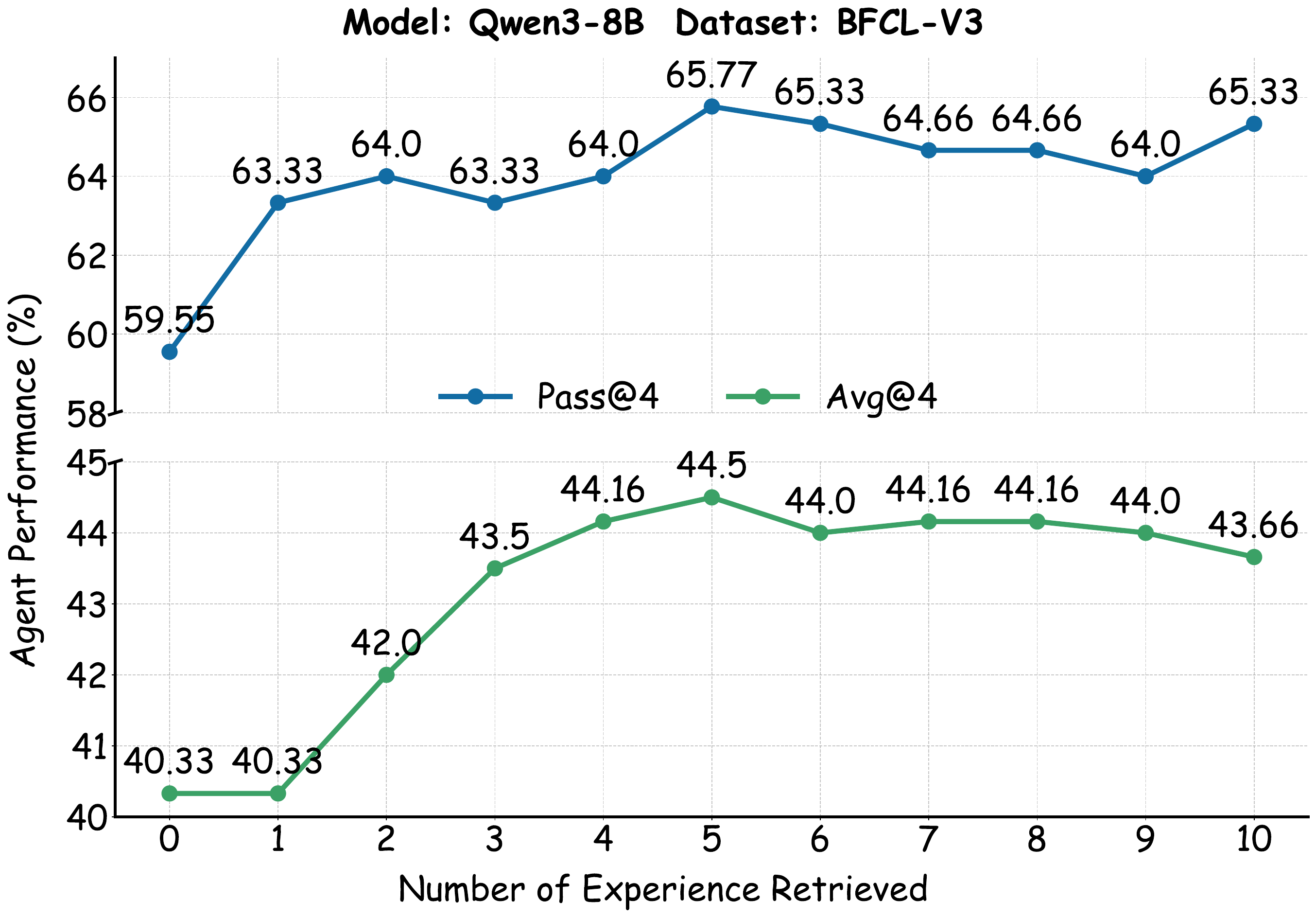}
    \caption{Effect of retrieved experience number on agent performance (\%) in {\ours} \texttt{(fixed)} setting.}
    \label{fig:ablation_on_k}
\end{figure}

\begin{table}[t]
\centering
\small
\begin{tabular}{llc}
    \toprule
    \textbf{Model} & \textbf{Setting} & \textbf{Latency (seconds)}\\
    \midrule
    \multirow{2}{*}{\textbf{Qwen3-8B}} & \textbf{No Memory} & 21.42 \\
    \cmidrule(l){2-3}
    & \textbf{+ {\ours}} & 23.96 $_{\textcolor{darkblue}{+2.54}}$  \\
    \bottomrule
\end{tabular}
\caption{The average inference latency per task for {\ours} versus the No Memory baseline.}\label{tab:latency}
\end{table}

\begin{figure}[t]
    \centering
    \includegraphics[width=1\linewidth]{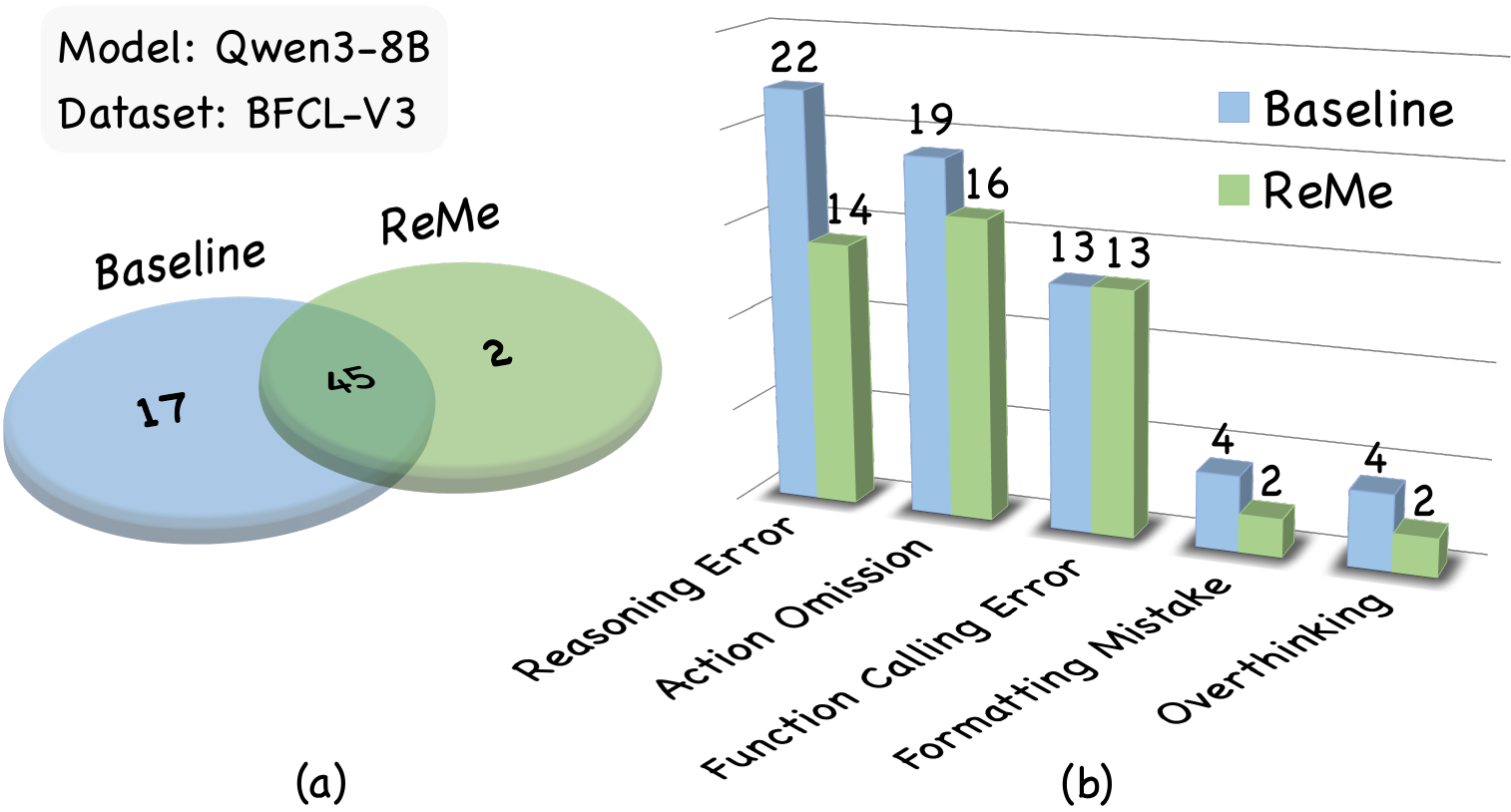}
    \caption{Statistics of failed tasks with and without {\ours}. (a) Left: shows overlapping and unique failure cases; (b) Right: displays the number of task failures across different error categories.}
    \label{fig:error_analysis}
\end{figure}

\subsection{More Analysis}\label{sec:analysis}
\paragraph{Agent Gains More with
Stronger $\textbf{LLM}_{\textbf{summ}}$.} 
Our main experiments demonstrate an agent can learn effectively through experience-driven self-evolution, i.e., $\text{LLM}_{summ}$=$\text{LLM}_{execute}$. 
To investigate whether the agent gains more as $\text{LLM}_{summ}$ capability increases, we scale the summarization model from Qwen3-8B to Qwen3-32B with the fixed $\text{LLM}_{execute}$ = Qwen3-8B. 
It can be observed from Table~\ref{tab:summarizer} that stronger summarization capability yields clear performance improvements in both Avg@4 and Pass@4 metrics (Avg@4: +1.83\% $\rightarrow$ +3.33\%; Pass@4: +0.23\% $\rightarrow$ +2.23\%). 
These findings emphasize the critical role of high-quality experience summarization in overall agent performance, highlighting the potential for further gains through advanced summarization techniques.

\paragraph{Effect of Retrieved Experience Number.} 
To evaluate the relationship between retrieved experience number and performance, we vary the value $K$ from 0 to 10. 
Figure~\ref{fig:ablation_on_k} illustrates increasing the number of in-context experiences achieves steady performance gains that rise and then saturate. 
Beyond the saturation point, retrieving more may degrade performance, primarily due to the higher chance of incorporating noisy experiences. 
This is why we select $K=5$ in the main experiments.

\paragraph{Computational Overheads.}
To demonstrate {\ours} offers a favorable trade-off between inference cost and performance enhancement, taking Appworld as an example, we report the average inference latency per task for {\ours} versus the No Memory baseline in Table~\ref{tab:latency}. 
It can be seen that the computational cost of our {\ours} is acceptable, which will not limit its applicability in long-running or resource-constrained settings.

\paragraph{Error Analysis.} 
We conduct an analysis of the error patterns with and without {\ours} for Qwen3-8B on BFCL-V3. 
The Venn diagram (Figure~\ref{fig:error_analysis}a) reveals a reduction in the total number of failure cases from 62 (No Memory Baseline) to 47 ({\ours}). 
Notably, {\ours} corrects 17 baseline-specific errors while introduces only 2 new ones. 
Further, we manually review and categorize each failure case to examine the impact of {\ours} on different error types (Figure~\ref{fig:error_analysis}b). 
A substantial decrease in \texttt{Reasoning Error} (22 $\rightarrow$ 14) indicates {\ours} effectively leverages past experiences to strengthen its multi‑step reasoning capabilities, reducing propagation of earlier mistakes. 
{\ours} also yields a moderate but meaningful reduction in \texttt{Action Omission} errors, which helps the agent recognize missing steps in multi-turn tasks, especially those requiring sequential tool interactions or state tracking.

\section{Conclusion}
We introduce {\ours}, a dynamic procedural memory framework that evolves agent reasoning from blind trial-and-error to strategic experience reuse. 
By distilling structured knowledge from prior trajectories at a fine-grained level, {\ours} enables agents to leverage critical insights, thus avoiding potential experience interference in coarse-grained approaches. 
Equipped with effective experience refinement, {\ours} maintains a high-quality experience pool for agent evolution. 
Extensive experiments validate that {\ours} significantly outperforms several baselines, with ablation studies highlighting the value of each core component in {\ours}. 

\section*{Limitations}

This paper focuses on procedural memory management for agent self-evolution. 
Despite its promising performance, there are several limitations that could be addressed in future work. 
First, {\ours} currently employs a fixed retrieval strategy, where experiences are retrieved once at the beginning of each task. 
Implementing a more flexible, context-aware retrieval mechanism could potentially improve system performance, since dynamic experience incorporation promotes adaptive knowledge utilization.
Secondly, although the existing experience validation process effectively filters out low-quality experiences, relying primarily on an LLM-as-judge approach may overlook nuanced aspects of experience quality and relevance. 
In the future, we can explore more sophisticated validation techniques for more precise experience evaluation. 
Furthermore, a larger-scale summarizer brings greater performance gains in agent reasoning, as shown in Section~\ref{sec:analysis}, which can be attributed to its stronger summarization capability. 
This indicates that designing advanced summarization strategies with small models can further boost agent self-evolution.


\bibliography{custom}

@article{tao2024survey,
  title={A survey on self-evolution of large language models},
  author={Tao, Zhengwei and Lin, Ting-En and Chen, Xiancai and Li, Hangyu and Wu, Yuchuan and Li, Yongbin and Jin, Zhi and Huang, Fei and Tao, Dacheng and Zhou, Jingren},
  journal={arXiv preprint arXiv:2404.14387},
  year={2024}
}

@article{gao2025survey,
  title={A survey of self-evolving agents: On path to artificial super intelligence},
  author={Gao, Huan-ang and Geng, Jiayi and Hua, Wenyue and Hu, Mengkang and Juan, Xinzhe and Liu, Hongzhang and Liu, Shilong and Qiu, Jiahao and Qi, Xuan and Wu, Yiran and others},
  journal={arXiv preprint arXiv:2507.21046},
  year={2025}
}

@article{fang2025comprehensive,
  title={A comprehensive survey of self-evolving ai agents: A new paradigm bridging foundation models and lifelong agentic systems},
  author={Fang, Jinyuan and Peng, Yanwen and Zhang, Xi and Wang, Yingxu and Yi, Xinhao and Zhang, Guibin and Xu, Yi and Wu, Bin and Liu, Siwei and Li, Zihao and others},
  journal={arXiv preprint arXiv:2508.07407},
  year={2025}
}

@article{zhang2025survey,
  title={A survey on the memory mechanism of large language model-based agents},
  author={Zhang, Zeyu and Dai, Quanyu and Bo, Xiaohe and Ma, Chen and Li, Rui and Chen, Xu and Zhu, Jieming and Dong, Zhenhua and Wen, Ji-Rong},
  journal={ACM Transactions on Information Systems},
  volume={43},
  number={6},
  pages={1--47},
  year={2025},
  publisher={ACM New York, NY}
}

@article{xu2025mem,
  title={A-mem: Agentic memory for llm agents},
  author={Xu, Wujiang and Mei, Kai and Gao, Hang and Tan, Juntao and Liang, Zujie and Zhang, Yongfeng},
  journal={arXiv preprint arXiv:2502.12110},
  year={2025}
}

@article{wang2025mirix,
  title={Mirix: Multi-agent memory system for llm-based agents},
  author={Wang, Yu and Chen, Xi},
  journal={arXiv preprint arXiv:2507.07957},
  year={2025}
}

@article{ding2024large,
  title={Large language model agent in financial trading: A survey},
  author={Ding, Han and Li, Yinheng and Wang, Junhao and Chen, Hang},
  journal={arXiv preprint arXiv:2408.06361},
  year={2024}
}

@article{wang2024large,
  title={Large language models for education: A survey and outlook},
  author={Wang, Shen and Xu, Tianlong and Li, Hang and Zhang, Chaoli and Liang, Joleen and Tang, Jiliang and Yu, Philip S and Wen, Qingsong},
  journal={arXiv preprint arXiv:2403.18105},
  year={2024}
}

@article{abbasian2023conversational,
  title={Conversational health agents: A personalized llm-powered agent framework},
  author={Abbasian, Mahyar and Azimi, Iman and Rahmani, Amir M and Jain, Ramesh},
  journal={arXiv preprint arXiv:2310.02374},
  year={2023}
}

@article{mei2025survey,
  title={A Survey of Context Engineering for Large Language Models},
  author={Mei, Lingrui and Yao, Jiayu and Ge, Yuyao and Wang, Yiwei and Bi, Baolong and Cai, Yujun and Liu, Jiazhi and Li, Mingyu and Li, Zhong-Zhi and Zhang, Duzhen and others},
  journal={arXiv preprint arXiv:2507.13334},
  year={2025}
}

@article{zhang2024survey,
  title={A survey on the memory mechanism of large language model based agents},
  author={Zhang, Zeyu and Dai, Quanyu and Bo, Xiaohe and Ma, Chen and Li, Rui and Chen, Xu and Zhu, Jieming and Dong, Zhenhua and Wen, Ji-Rong},
  journal={ACM Transactions on Information Systems},
  year={2024},
  publisher={ACM New York, NY}
}

@inproceedings{
  wang2025agent,
  title={Agent Workflow Memory},
  author={Zora Zhiruo Wang and Jiayuan Mao and Daniel Fried and Graham Neubig},
  booktitle={Forty-second International Conference on Machine Learning},
  year={2025}
}

@article{qiao2024agent,
  title={Agent planning with world knowledge model},
  author={Qiao, Shuofei and Fang, Runnan and Zhang, Ningyu and Zhu, Yuqi and Chen, Xiang and Deng, Shumin and Jiang, Yong and Xie, Pengjun and Huang, Fei and Chen, Huajun},
  journal={Advances in Neural Information Processing Systems},
  volume={37},
  pages={114843--114871},
  year={2024}
}

@article{ganguli2025mark,
  title={MARK: Memory Augmented Refinement of Knowledge},
  author={Ganguli, Anish and Deb, Prabal and Banerjee, Debleena},
  journal={arXiv preprint arXiv:2505.05177},
  year={2025}
}

@inproceedings{zhao2024expel,
  title={Expel: Llm agents are experiential learners},
  author={Zhao, Andrew and Huang, Daniel and Xu, Quentin and Lin, Matthieu and Liu, Yong-Jin and Huang, Gao},
  booktitle={Proceedings of the AAAI Conference on Artificial Intelligence},
  volume={38},
  pages={19632--19642},
  year={2024}
}

@inproceedings{tan-etal-2025-prospect,
    title = "In Prospect and Retrospect: Reflective Memory Management for Long-term Personalized Dialogue Agents",
    author = "Tan, Zhen  and Yan, Jun  and Hsu, I-Hung  and Han, Rujun  and Wang, Zifeng  and Le, Long  and Song, Yiwen  and Chen, Yanfei  and Palangi, Hamid  and Lee, George  and Iyer, Anand Rajan  and Chen, Tianlong  and Liu, Huan  and Lee, Chen-Yu  and Pfister, Tomas",
    booktitle = "Proceedings of the 63rd Annual Meeting of the Association for Computational Linguistics (Volume 1: Long Papers)",
    year = "2025",
    pages = "8416--8439"
}

@inproceedings{zheng2023synapse,
  title={Synapse: Trajectory-as-Exemplar Prompting with Memory for Computer Control},
  author={Zheng, Longtao and Wang, Rundong and Wang, Xinrun and An, Bo},
  booktitle={The Twelfth International Conference on Learning Representations},
  year={2024}
}

@article{hu2024hiagent,
  title={Hiagent: Hierarchical working memory management for solving long-horizon agent tasks with large language model},
  author={Hu, Mengkang and Chen, Tianxing and Chen, Qiguang and Mu, Yao and Shao, Wenqi and Luo, Ping},
  journal={arXiv preprint arXiv:2408.09559},
  year={2024}
}

@article{tang2025agentkb,
  title={Agent KB: Leveraging Cross-Domain Experience for Agentic Problem Solving},
  author={Tang, Xiangru and Qin, Tianrui and Peng, Tianhao and Zhou, Ziyang and Shao, Daniel and Du, Tingting and Wei, Xinming and Xia, Peng and Wu, Fang and Zhu, He and others},
  journal={arXiv preprint arXiv:2507.06229},
  year={2025}
}

@article{chen2025swe,
  title={SWE-Exp: Experience-Driven Software Issue Resolution},
  author={Chen, Silin and Lin, Shaoxin and Gu, Xiaodong and Shi, Yuling and Lian, Heng and Yun, Longfei and Chen, Dong and Sun, Weiguo and Cao, Lin and Wang, Qianxiang},
  journal={arXiv preprint arXiv:2507.23361},
  year={2025}
}

@inproceedings{liu-etal-2025-contextual,
    title = "Contextual Experience Replay for Self-Improvement of Language Agents",
    author = "Liu, Yitao  and Si, Chenglei  and Narasimhan, Karthik R  and Yao, Shunyu",
    booktitle = "Proceedings of the 63rd Annual Meeting of the Association for Computational Linguistics (Volume 1: Long Papers)",
    year = "2025",
    pages = "14179--14198"
}

@article{xiong2025memory,
  title={How Memory Management Impacts LLM Agents: An Empirical Study of Experience-Following Behavior},
  author={Xiong, Zidi and Lin, Yuping and Xie, Wenya and He, Pengfei and Tang, Jiliang and Lakkaraju, Himabindu and Xiang, Zhen},
  journal={arXiv preprint arXiv:2505.16067},
  year={2025}
}

@article{zheng2023judging,
  title={Judging llm-as-a-judge with mt-bench and chatbot arena},
  author={Zheng, Lianmin and Chiang, Wei-Lin and Sheng, Ying and Zhuang, Siyuan and Wu, Zhanghao and Zhuang, Yonghao and Lin, Zi and Li, Zhuohan and Li, Dacheng and Xing, Eric and others},
  journal={Advances in neural information processing systems},
  volume={36},
  pages={46595--46623},
  year={2023}
}

@inproceedings{patilberkeley,
  title={The Berkeley Function Calling Leaderboard (BFCL): From Tool Use to Agentic Evaluation of Large Language Models},
  author={Patil, Shishir G and Mao, Huanzhi and Yan, Fanjia and Ji, Charlie Cheng-Jie and Suresh, Vishnu and Stoica, Ion and Gonzalez, Joseph E},
  booktitle={Forty-second International Conference on Machine Learning},
  year={2025}
}

@inproceedings{trivedi2024appworld,
  title={AppWorld: A Controllable World of Apps and People for Benchmarking Interactive Coding Agents},
  author={Trivedi, Harsh and Khot, Tushar and Hartmann, Mareike and Manku, Ruskin and Dong, Vinty and Li, Edward and Gupta, Shashank and Sabharwal, Ashish and Balasubramanian, Niranjan},
  booktitle={Proceedings of the 62nd Annual Meeting of the Association for Computational Linguistics (Volume 1: Long Papers)},
  pages={16022--16076},
  year={2024}
}

@misc{LangChain2025,
  author = {{LangChain}},
  title = {LangMem: Modular memory for agentic systems},
  year = {2025},
  url = {https://github.com/langchain-ai/langmem},
  note = {Accessed: 2025-10-13},
  howpublished = {\url{https://github.com/langchain-ai/langmem}}
}

@inproceedings{
    shinn2023reflexion,
    title={Reflexion: language agents with verbal reinforcement learning},
    author={Noah Shinn and Federico Cassano and Ashwin Gopinath and Karthik R Narasimhan and Shunyu Yao},
    booktitle={Thirty-seventh Conference on Neural Information Processing Systems},
    year={2023}
}

@misc{qwen3technicalreport,
      title={Qwen3 Technical Report}, 
      author={Qwen Team},
      year={2025},
      eprint={2505.09388},
      archivePrefix={arXiv},
      primaryClass={cs.CL},
      url={https://arxiv.org/abs/2505.09388}, 
}

@article{qwen3embedding,
  title={Qwen3 Embedding: Advancing Text Embedding and Reranking Through Foundation Models},
  author={Zhang, Yanzhao and Li, Mingxin and Long, Dingkun and Zhang, Xin and Lin, Huan and Yang, Baosong and Xie, Pengjun and Yang, An and Liu, Dayiheng and Lin, Junyang and Huang, Fei and Zhou, Jingren},
  journal={arXiv preprint arXiv:2506.05176},
  year={2025}
}

@misc{gpt-4.1,
  author = {OpenAI},
  title = {Introducing GPT-4.1 in the API},
  year = {2025},
  url = {https://openai.com/index/gpt-4-1/}
}

@misc{o4-mini,
  author = {OpenAI},
  title = {Introducing OpenAI o3 and o4-mini},
  year = {2025},
  url = {https://openai.com/index/introducing-o3-and-o4-mini/}
}

@misc{qwen3max,
    title = {Qwen3-Max: Just Scale it},
    author = {Qwen Team},
    month = {September},
    year = {2025}
}

@misc{kimi-k2-thinking,
  author = {MoonshotAI},
  title = {Introducing Kimi K2 Thinking},
  year = {2025},
  url = {https://moonshotai.github.io/Kimi-K2/thinking.html}
}

@article{liu2025deepseek,
  title={Deepseek-v3. 2: Pushing the frontier of open large language models},
  author={Liu, Aixin and Mei, Aoxue and Lin, Bangcai and Xue, Bing and Wang, Bingxuan and Xu, Bingzheng and Wu, Bochao and Zhang, Bowei and Lin, Chaofan and Dong, Chen and others},
  journal={arXiv preprint arXiv:2512.02556},
  year={2025}
}

@misc{glm-4.7,
  author = {ZhipuAI},
  title = {GLM-4.7: Advancing the Coding Capability},
  year = {2025},
  url = {https://z.ai/blog/glm-4.7}
}

\appendix
\section{Use of AI Assistants}
During the writing process of this paper, we utilized AI assistants for language refinement, including tasks such as grammar checking, sentence restructuring, and phrasing. 
All AI-suggested changes were thoroughly reviewed and approved by the authors to ensure the final content of this paper represents the authors' original ideas. 

\section{Experimental Details}
In this section, we detail our experimental setup, including the benchmark datasets, evaluation metrics, and baseline methods compared. We also describe the implementation details of {\ours} crucial for reproducing our results.

\subsection{Datasets and Evaluation Metrics}\label{app:dataset}
\paragraph{BFCL-V3} Berkeley Function Calling Leaderboard V3 (BFCL-V3)~\citep{patilberkeley} is a benchmark which assesses the function calling and tool-using capabilities of LLMs, particularly in multi-turn and multi-step scenarios. 
It provides over 1,800 test tasks that require models to generate precise API calls, handle various programming languages (Python, Java, JavaScript), and manage complex interactions like parallel function calls. 
The evaluation employs both Abstract Syntax Tree (AST) matching to check syntactic correctness and executable testing to verify functional outcomes. 
In our experiments, a task is deemed successful when the agent makes the necessary function calls correctly and yields the expected outputs. 

\paragraph{AppWorld} AppWorld~\citep{trivedi2024appworld} is a benchmark designed to evaluate function calling and interactive coding agents. 
It simulates a world of 9 day-to-day applications (e.g., email, Spotify, Venmo) through 457 APIs and is populated with the digital activities of approximately 100 simulated users.
A key feature of AppWorld is its robust evaluation framework, which uses state-based unit tests to assess task completion and provides two metrics to measure performance: 1) Task Goal Completion (TGC) measures percentage of tasks for which the agent passes all evaluation tests; 2) Scenario Goal Completion (SGC) is the percentage of scenarios where the agent passes all the unit tests for all tasks from that scenario.
In our experiments, we report Task Goal Completion metric, which naturally reflects task success rate.

\begin{figure}[tbp]
    \centering
    \includegraphics[width=1\linewidth]{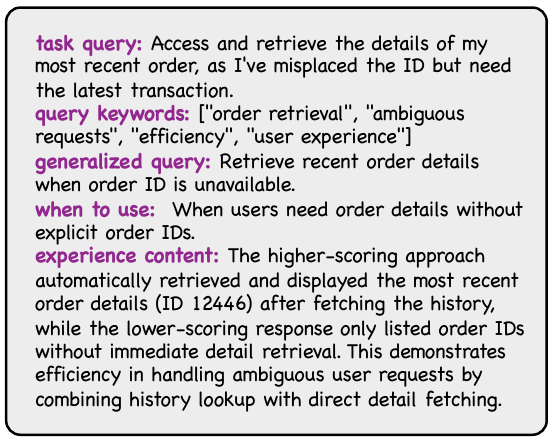}
    \caption{Different indexing examples for the same BFCL-V3 task experience.}
    \label{fig:indexing_example}
\end{figure}

\subsection{Baseline Details}\label{app:baselines}
\paragraph{LangMem} LangMem~\citep{LangChain2025} is Langchain’s long-term memory module that extracts and stores key information from conversations for future retrieval. 
It provides both functional primitives compatible with any storage system and native integration with LangGraph's storage layer, enabling agents to continuously improve. 
In our experiments, we adopt LangMem’s implementation of episodic memory\footnote{https://langchain-ai.github.io/langmem/guides/extract\_ episodic\_memories/}, which helps the agent learn from experience.

\paragraph{A-Mem} A-Mem~\citep{xu2025mem} is a system designed to provide LLM agents with agentic memory, allowing them to autonomously manage their own long-term knowledge. 
It constructs a memory-centric knowledge graph for agents, actively deciding what information to store, recall, and update based on their goals and interaction.
In our experiments, we reproduce A-Mem using its open-source code, with slight prompt modifications to extract procedural memories.

\begin{figure}[t]
    \centering
    \includegraphics[width=1\linewidth]{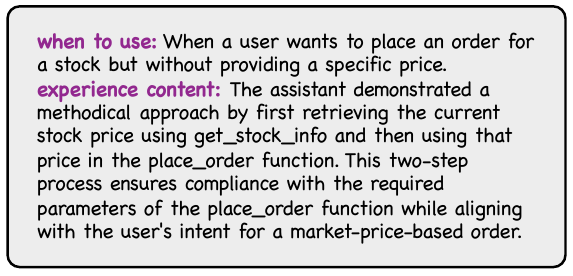}
    \caption{Experience example on BFCL-V3.}
    \label{fig:exp_example_bfcl}
\end{figure}
\begin{figure}[t]
    \centering
    \includegraphics[width=1\linewidth]{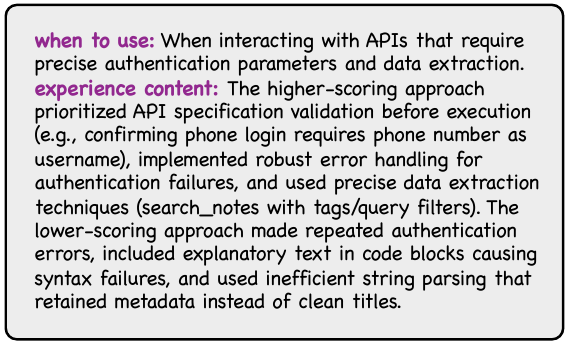}
    \caption{Experience example on AppWorld.}
    \label{fig:exp_example_app}
\end{figure}
\begin{figure*}[t]
    \centering
    \includegraphics[width=1\linewidth]{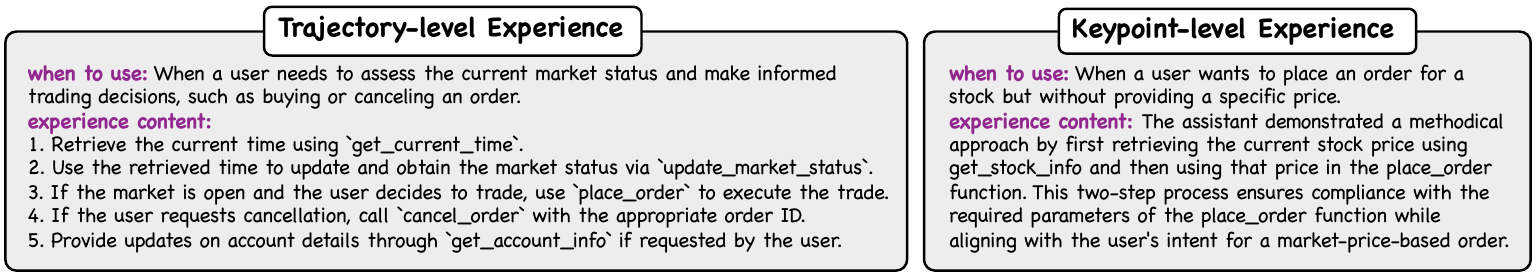}
    \caption{Comparison of trajectory-level and keypoint-level experience granularity.}
    \label{fig:granularity}
\end{figure*}

\subsection{Model Configuration}
We select Qwen3 series for backbone models in our main experiments. By using Qwen3-Instruct models of different sizes, including 8B, 14B, and 32B, we can explore whether memory-enhanced agent performance improves as model size increases. For high-quality experience acquisition, Qwen3 thinking mode is activated for BFCL-V3 tasks and disabled for AppWorld tasks. 

To validate the generalizability of our system, we also evaluate other LLMs with and without {\ours} on BFCL-V3 tasks, including GPT‑4.1-2025-04-14~\cite{gpt-4.1}, o4‑mini-2025-04-16~\cite{o4-mini}, Qwen3‑Max-Preview~\cite{qwen3max}, Kimi‑K2‑Thinking~\cite{kimi-k2-thinking}, DeepSeek‑V3.2~\cite{liu2025deepseek}, and GLM‑4.7~\cite{glm-4.7}, where Qwen3‑Max‑Preview, DeepSeek‑V3.2, and Kimi‑K2‑Thinking are in the thinking mode, and the others are in the non‑thinking mode.

\subsection{Implementation Details}
\subsubsection{For Experience Acquisition}\label{app:settings}
First, we sample trajectories $N$= 8 times for each task query to obtain a diverse set of potential solutions including both high-reward and low-reward results. 
Next, within each group corresponding to the same task, all trajectories are sorted by their rewards and only the lowest‑scoring and highest‑scoring examples are selected to the following experience acquisition. 

\begin{itemize}[leftmargin=12pt,topsep=2pt,itemsep=-2pt]
    \item \textbf{Success Pattern Recognition}: Successful trajectories are defined as those exceeding a predefined score threshold (empirically set to 1.0). Then, we prompt $\text{LLM}_{summ}$ to identify the key point that contributes to the task success. 
    \item \textbf{Failure Analysis}: Conversely, failed trajectories trigger failure analysis by prompting $\text{LLM}_{summ}$ to determine the earliest key step that leads to suboptimal outcomes.
    \item \textbf{Comparative Insight Generation}: When the reward gap exists between the chosen two trajectories, we prompt $\text{LLM}_{summ}$ to articulate which specific decision or action distinguishes higher‑scoring from lower‑scoring attempts. 
\end{itemize}

Three example prompts for experience acquisition are shown in Table ~\ref{tab:success_prompt},~\ref{tab:failure_prompt} and~\ref{tab:compare_prompt}. 
To filter out the generated invalid experiences, we employ the LLM-as-a-Judge prompt in Table~\ref{tab:validation_prompt} for validation.

\begin{table}[tbp]
\centering
\setlength\tabcolsep{2pt}
\renewcommand\arraystretch{1.08}
\resizebox{\linewidth}{!}{
    \begin{tabular}{@{}lccc@{}}
    \toprule
    \textbf{Model} & \textbf{No Memory} & \textbf{{\ours} \texttt{(fixed)}} & \textbf{{\ours} \texttt{(dynamic)}}\\ 
    \midrule
    \textbf{GPT-4.1} & 48.25\% & 53.33\%$_{\textcolor{darkblue}{+5.08\%}}$&54.67\%$_{\textcolor{darkblue}{+6.42\%}}$\\ 
    \textbf{o4-mini} & 54.67\% & 57.78\%$_{\textcolor{darkblue}{+3.11\%}}$&60.00\%$_{\textcolor{darkblue}{+5.33\%}}$ \\
    \textbf{Qwen3-Max} &59.00\% &61.83\%$_{\textcolor{darkblue}{+2.83\%}}$&64.00\%$_{\textcolor{darkblue}{+5.00\%}}$\\
    \textbf{Kimi-K2-Thinking} & 57.17\%&62.17\%$_{\textcolor{darkblue}{+5.00\%}}$&66.00\%$_{\textcolor{darkblue}{+8.83\%}}$ \\
    \textbf{DeepSeek-V3.2} & 53.06\%& 56.00\%$_{\textcolor{darkblue}{+2.94\%}}$& 57.33\%$_{\textcolor{darkblue}{+4.27\%}}$\\
    \textbf{GLM-4.7} &68.00\% &70.00\%$_{\textcolor{darkblue}{+2.00\%}}$&73.83\%$_{\textcolor{darkblue}{+5.83\%}}$\\
    \bottomrule
    \end{tabular}
}
\caption{Performance comparison of more LLMs with and without {\ours} on BFCL-V3 (Avg@4 \%).}\label{tab:more_models}
\end{table}

\begin{table*}[tbp]
\centering
\setlength\tabcolsep{8pt}
\resizebox{\linewidth}{!}{
    \begin{tabular}{cccccc}
    \toprule
    \textbf{BFCL-V3} & \textbf{Overall (750)} &	\textbf{Base (150)}	& \textbf{Miss Param (200)} & \textbf{Miss Func (200)}	&\textbf{Long Context (200)}\\
    \midrule
    \textbf{Qwen3-8B} & 37.78\% & 59.55\% & 31.00\% & 19.00\% & 47.00\% \\
    \rowcolor{lightblue!20} \textbf{+{\ours}} & 43.02\%$_{\textcolor{red}{+5.24\%}}$	& 65.77\%$_{\textcolor{red}{+6.22\%}}$ & 38.50\%$_{\textcolor{red}{+7.50\%}}$	& 21.50\%$_{\textcolor{red}{+2.50\%}}$ & 52.00\%$_{\textcolor{red}{+5.00\%}}$\\
    \bottomrule
    \end{tabular}
}
\caption{Evaluation results on complete BFCL V3 Multi-Turn category.}\label{tab:larger}
\end{table*}

\begin{table}[tbp]
    \centering
    \resizebox{\linewidth}{!}{
    \begin{tabular}{llcc}
    \toprule
        \textbf{Model} & \textbf{Retrieval Key} & \textbf{Avg@4} & \textbf{Pass@4}\\ 
    \midrule
        \multirow{4}{*}{\textbf{Qwen3-8B}} & \textit{task query} & 44.00\%&63.33\% \\ 
         & \textit{generalized query}& 42.50\%&63.77\%\\ 
         & \textit{query keywords} & 44.22\% & 65.33\% \\ 
         & \textit{usage scenario} & \textbf{44.50\%} & \textbf{65.77\%} \\ 
    \midrule
        \multirow{4}{*}{\textbf{Qwen3-14B}} & \textit{task query} & 50.11\%& 71.77\% \\ 
         & \textit{generalized query} &50.49\%&72.22\% \\ 
         & \textit{query keywords} & 51.16\%&71.11\% \\ 
         & \textit{usage scenario}  & \textbf{51.89\%} & \textbf{72.44\%} \\ 
    \midrule
        \multirow{4}{*}{\textbf{Qwen3-32B}} & \textit{task query} & 56.22\%	&72.22\% \\ 
         & \textit{generalized query} &55.33\%&73.33\% \\ 
         & \textit{query keywords} &  \textbf{56.89\%}	&74.44\%\\ 
         & \textit{usage scenario}  & 56.05\% & \textbf{74.89\%} \\ 
    \bottomrule
    \end{tabular}
    }
    \caption{Ablation study of retrieve keys.}\label{tab:key}
\end{table}

\subsubsection{For Experience Retrieval}\label{app:retrieval}
When a new task is received, $\text{LLM}_{execute}$ retrieves relevant experiences $\mathcal{E}_{r}$ by matching the current task’s query $q_{new}$ against the usage scenario field $w$ of stored experiences:
\begin{equation}
    \mathcal{E}_{r} = \arg{\text{top}_k} \left[sim_{cos}(E_i, q_{new})\right].
\end{equation}
Here, $sim_{cos}$ stands for the computation of cosine similarity between embeddings. 
In our experiments, past experiences are indexed using vector representations of the usage scenario field $\phi(w)$, obtained from Qwen3-Embedding model $\phi(\cdot)$.
Our selected vector database is Elasticsearch.
\begin{equation}
    sim_{cos}(E, q_{new})= \frac{\phi(w)\cdot \phi(q_{new}) }{\|\phi(w)\| \, \|\phi(q_{new})\|}
\end{equation}

In Section~\ref{sec:ablation}, we also explore more indexing strategies for experience storage. 
The example in Figure~\ref{fig:indexing_example} illustrates the differences among these retrieval keys. 

\section{Experience Examples}\label{app:example}
{\ours} focuses on extracting keypoint-level experiences from historical trajectories, with examples for BFCL-V3 and AppWorld illustrated in Figure~\ref{fig:exp_example_bfcl} and~\ref{fig:exp_example_app}, respectively. 
To further investigate the impact of experience granularity, we compare trajectory-level and keypoint-level acquisition, as described in Section~\ref{sec:ablation}. 
In Figure~\ref{fig:granularity}, we contrast the structural and content characteristics of the two granularity levels, showing how trajectory-level captures exhaustive procedural details, while keypoint-level emphasizes critical actions and omits less relevant steps.

\section{Detailed Experimental Results}
\subsection{Generalizability across Different Models}
To examine the generalizability of {\ours}, we conduct experiments with different backbone models, including GPT-4.1, o4‑mini, Qwen3‑Max, Kimi‑K2, DeepSeek‑V3.2, and GLM‑4.7. 
The stacked analysis in Figure~\ref{fig:more_models} and detailed results in Table~\ref{tab:more_models} show that {\ours} consistently outperforms No Memory baseline across various backbones, confirming the model‑agnostic advantage of our novel procedural memory system.

\subsection{Results on Larger-scale Benchmark}\label{app:larger_benchmark}
To further show the superiority of {\ours}, we supplement the experiments on the Missing Parameters, Missing Functions, and Long Context category of BFCL-V3 benchmark with total 600 tasks. 
Using the experience pool constructed from 50 tasks in base multi-turn category, the evaluation results (Pass@4 as metric) are shown in Table~\ref{tab:larger}.
It can be observed that our {\ours} still performs well when evaluated on a larger-size benchmark.

\subsection{Retrieval Key Analysis}\label{app:retrieval_key}
Table~\ref{tab:key} compares four retrieval key strategies (task query, generalized query, query keywords, and usage scenario) across three model scales (Qwen3–8B, Qwen3–14B, and Qwen3–32B) on the BFCL-V3 benchmark under the {\ours}\texttt{(fixed)} setting. 
Consistent with the trends observed in Figure~\ref{fig:key}, simple indexing methods such as raw task query and query keywords generally yield lower performance. 
In contrast, LLM-generated retrieval keys, particularly the usage scenario field, exhibit consistently strong results across all models, achieving the highest or near-highest Avg@4 and Pass@4 scores.

\section{Case Study}
To gain deeper insights into how experience reuse influences agent reasoning, we compare two agent trajectories on the same BFCL-V3 task, one guided by retrieved experiences and one without. 
As illustrated in Figure~\ref{fig:example}, without past experience, the agent encounters a failure when purchasing Apple shares since it fabricates the current market price instead of fetching real-time data. 
With {\ours}, past experience guides the agent to correctly obtain real-time pricing before placing an order, successfully completing the stock trading task. 
This case demonstrates how experience-driven reasoning prevents agents from repeating earlier mistakes and improves robustness across similar scenarios.

\begin{table*}
\centering
\begin{tabular}{c}
\begin{tcolorbox}[title={Example Prompt for Success Pattern Recognition},breakable,]
  You are an expert AI analyst reviewing successful step sequences from an AI agent execution.\\
  
  Your task is to extract reusable, actionable step-level experiences that can guide future agent executions.\\
  Focus on identifying specific patterns, techniques, and decision points that contributed to success.\\
  
  ANALYSIS FRAMEWORK:\\
  $\bullet$ STEP PATTERN ANALYSIS: Identify the specific sequence of actions that led to success\\
  $\bullet$ DECISION POINTS: Highlight critical decisions made during these steps\\
  $\bullet$ TECHNIQUE EFFECTIVENESS: Analyze why specific approaches worked well\\
  $\bullet$ REUSABILITY: Extract patterns that can be applied to similar scenarios\\
  
  EXTRACTION PRINCIPLES:\\
  $\bullet$ Focus on TRANSFERABLE TECHNIQUES and decision frameworks\\
  $\bullet$ Frame insights as actionable guidelines and best practices\\
  
  \# Original Query\\
  \{query\}\\
  
  \# Step Sequence Analysis\\
  \{step\_sequence\}\\
  
  \# Context Information\\
  \{context\}\\
  
  \# Outcome\\
  This step sequence was part of a successful trajectory.\\
  
  OUTPUT FORMAT:\\
  Generate 1-3 step-level success insights as JSON objects:\\
  \verb|```|json\\
  $\left[ \right.\\
  \left. \right. \; \left\{ \left\{ \right.\right.\\
      \left. \right. \quad``\text{when\_to\_use}": ``\text{Specific conditions when this success insight should be applied}",\\
      \left. \right. \quad``\text{task\_query}": ``\text{Identify the specific task query from the original trajectory that this success}\\ \text{experience is most closely related to. Extract the exact query text.}",\\
      \left. \right. \quad``\text{generalized\_query}": ``\text{Abstract the specific task query to create a more generalized task} \\ \text{representation.}",\\
      \left. \right. \quad ``\text{experience}": ``\text{Detailed description of the successful step pattern and why it works}",\\
      \left. \right. \quad ``\text{tags}": [\text{``relevant", ``keywords", ``from", ``the", ``task", ``query"}],\\
      \left. \right. \quad ``\text{confidence}": 0.8,\\
      \left. \right. \quad ``\text{tools\_used}": [\text{``list", ``of", ``tools"}]\\
  \left. \right. \; \left. \left. \right\} \right\}\\
  \left. \right]$\\
  \verb|```|
\end{tcolorbox}
\end{tabular}
\caption{Example prompt for success pattern recognition.}
\label{tab:success_prompt}
\end{table*}

\begin{table*}
\centering
\begin{tabular}{c}
\begin{tcolorbox}[title={Example Prompt for Failure Analysis},breakable,]
  You are an expert AI analyst reviewing failed step sequences from an AI agent execution.\\
  
  Your task is to extract learning experiences from failures to prevent similar mistakes in future executions.\\
  Focus on identifying error patterns, missed opportunities, and alternative approaches.\\
  
  ANALYSIS FRAMEWORK:\\
  $\bullet$ FAILURE POINT IDENTIFICATION: Pinpoint where and why the steps went wrong\\
  $\bullet$ ERROR PATTERN ANALYSIS: Identify recurring mistakes or problematic approaches\\
  $\bullet$ ALTERNATIVE APPROACHES: Suggest what could have been done differently\\
  $\bullet$ PREVENTION STRATEGIES: Extract actionable insights to avoid similar failures\\
  
  EXTRACTION PRINCIPLES:\\
  $\bullet$ Extract GENERAL PRINCIPLES as well as SPECIFIC INSTRUCTIONS\\
  $\bullet$ Focus on PATTERNS and RULES as well as particular instances\\
  
  \# Original Query\\
  \{query\}\\
  
  \# Step Sequence Analysis\\
  \{step\_sequence\}\\
  
  \# Context Information\\
  \{context\}\\
  
  \# Outcome\\
  This step sequence was part of a failed trajectory.\\
  
  OUTPUT FORMAT:\\
  Generate 1-3 step-level failure prevention insights as JSON objects:\\
  \verb|```|json\\
  $\left[ \right.\\
  \left. \right. \; \left\{ \left\{ \right.\right.\\
      \left. \right. \quad``\text{when\_to\_use}": ``\text{Specific situations where this lesson should be remembered}",\\
      \left. \right. \quad``\text{task\_query}": ``\text{Identify the specific task query from the original trajectory that this lesson is}\\ \text{most closely related to. Extract the exact query text.}",\\
      \left. \right. \quad``\text{generalized\_query}": ``\text{Abstract the specific task query to create a more generalized task } \\ \text{representation.}",\\
      \left. \right. \quad ``\text{experience}": ``\text{Universal principle or rule extracted from the failure pattern}",\\
      \left. \right. \quad ``\text{tags}": [\text{``relevant", ``keywords", ``from", ``the", ``task", ``query"}],\\
      \left. \right. \quad ``\text{confidence}": 0.8,\\
      \left. \right. \quad ``\text{tools\_used}": [\text{``list", ``of", ``tools"}]\\
  \left. \right. \; \left. \left. \right\} \right\}\\
  \left. \right]$\\
  \verb|```|
\end{tcolorbox}
\end{tabular}
\caption{Example prompt for failure analysis.}
\label{tab:failure_prompt}
\end{table*}

\begin{table*}
\centering
\begin{tabular}{c}
\begin{tcolorbox}[title={Example Prompt for Comparative Insights Generation},breakable,]
   You are an expert AI analyst comparing higher-scoring and lower-scoring step sequences to extract performance insights.\\
  
  Your task is to identify the key differences between higher and lower performing approaches at the step level.\\
  Focus on what made the higher-scoring approach more effective, even when both approaches may have had partial success.\\
  
  SOFT COMPARATIVE ANALYSIS FRAMEWORK:\\
  $\bullet$ PERFORMANCE FACTORS: Identify what specifically contributed to the higher score\\
  $\bullet$ APPROACH DIFFERENCES: Compare methodologies and execution strategies\\
  $\bullet$ EFFICIENCY ANALYSIS: Analyze why one approach was more efficient or effective\\
  $\bullet$ OPTIMIZATION INSIGHTS: Extract lessons for improving performance\\
  
  EXTRACTION PRINCIPLES:\\
  $\bullet$ Focus on INCREMENTAL IMPROVEMENTS and performance optimization\\
  $\bullet$ Extract QUALITY INDICATORS that differentiate better vs good approaches\\
  $\bullet$ Identify REFINEMENT STRATEGIES that lead to higher scores\\
  $\bullet$ Frame insights as PERFORMANCE ENHANCEMENT guidelines \\
  
  \# Higher-Scoring Step Sequence (Score: \{higher\_score\}) \\
  \{higher\_steps\}\\
  
  \# Lower-Scoring Step Sequence (Score: \{lower\_score\})\\
  \{lower\_steps\}\\
  
  OUTPUT FORMAT:\\
  Generate 1-2 performance improvement insights as JSON objects:\\
  \verb|```|json\\
  $\left[ \right.\\
  \left. \right. \; \left\{ \left\{ \right.\right.\\
      \left. \right. \quad``\text{when\_to\_use}": ``\text{Specific scenarios where this performance insight applies}",\\
      \left. \right. \quad``\text{task\_query}": ``\text{Identify the specific task query from the original trajectory that this }\\ \text{performance insight is most closely related to. Extract the exact query text.}",\\
      \left. \right. \quad``\text{generalized\_query}": ``\text{Abstract the specific task query to create a more generalized task } \\ \text{representation.}",\\
      \left. \right. \quad ``\text{experience}": ``\text{Detailed analysis of what made the higher-scoring approach more effective}",\\
      \left. \right. \quad ``\text{tags}": [\text{``relevant", ``keywords", ``from", ``the", ``task", ``query"}],\\
      \left. \right. \quad ``\text{confidence}": 0.8,\\
      \left. \right. \quad ``\text{tools\_used}": [\text{``list", ``of", ``tools"}]\\
  \left. \right. \; \left. \left. \right\} \right\}\\
  \left. \right]$\\
  \verb|```|
\end{tcolorbox}
\end{tabular}
\caption{Example prompt for comparative insights generation.}
\label{tab:compare_prompt}
\end{table*}

\begin{table*}
\centering
\begin{tabular}{c}
\begin{tcolorbox}[title={Example Prompt for Experience Validation},breakable,]
  You are an expert AI analyst tasked with validating the quality and usefulness of extracted step-level experiences.\\
  
  Your task is to assess whether the extracted experience is actionable, accurate, and valuable for future agent executions.\\
  
  VALIDATION CRITERIA:\\
  $\bullet$ ACTIONABILITY: Is the experience specific enough to guide future actions?\\
  $\bullet$ ACCURACY: Does the experience correctly reflect the patterns observed?\\
  $\bullet$ RELEVANCE: Is the experience applicable to similar future scenarios?\\
  $\bullet$ CLARITY: Is the experience clearly articulated and understandable?\\
  $\bullet$ UNIQUENESS: Does the experience provide novel insights or common knowledge?\\
  
  \# Experience to Validate\\
  Condition: {condition}\\
  Experience Content: {experience\_content}\\
  
  OUTPUT FORMAT:\\
  Provide validation assessment:\\
  \verb|```|json\\
  $\left\{ \left\{ \right.\right.\\
    \left. \right. \; ``\text{is\_valid}": \text{true/false},\\
    \left. \right. \; ``\text{score}": 0.8,\\
    \left. \right. \; ``\text{feedback}": ``\text{Detailed explanation of validation decision}",\\
    \left. \right. \; ``\text{recommendations}": ``\text{Suggestions for improvement if applicable}"\\
  \left. \left. \right\} \right\}$\\
  \verb|```|
  
  Score should be between 0.0 (poor quality) and 1.0 (excellent quality).\\
  Mark as invalid if score is below 0.3 or if there are fundamental issues with the experience.\\
  
\end{tcolorbox}
\end{tabular}
\caption{Example prompt for experience validation.}
\label{tab:validation_prompt}
\end{table*}

\begin{table*}
\centering
\begin{tabular}{c}
\begin{tcolorbox}[title={Example Prompt for Experience Reranking},breakable,]
  You are an expert AI analyst tasked with reranking retrieved experiences based on their relevance to a specific query.\\
  
  Your task is to analyze the candidates and rank them by relevance, considering:\\
  $\bullet$ DIRECT RELEVANCE: How directly applicable the experience is to the current query\\
  $\bullet$ SITUATION SIMILARITY: How similar the experience context is to the current situation\\
  $\bullet$ ACTIONABILITY: How actionable and specific the experience is\\
  $\bullet$ QUALITY: The overall quality and clarity of the experience\\
  
  \# Current Query\\
  {query}\\
  
  \# Candidate Experiences (Total: {num\_candidates})\\
  {candidates}\\
  
  OUTPUT FORMAT:\\
  Provide a ranked list of candidate indices (0-based) from most relevant to least relevant:\\
  \verb|```|json\\
  $\left\{ \left\{ \right.\right.\\
  \left. \right. \; ``\text{ranked\_indices}": [2, 0, 4, 1, 3],\\
  \left. \right. \; ``\text{reasoning}": ``\text{Brief explanation of ranking rationale}"\\
  \left. \left. \right\} \right\}$\\
  \verb|```|\\
  
  Note: Include ALL candidate indices in the ranking, even if some are less relevant.\\
\end{tcolorbox}
\end{tabular}
\caption{Example prompt for experience reranking.}
\label{tab:rerank_prompt}
\end{table*}

\begin{table*}
\centering
\begin{tabular}{c}
\begin{tcolorbox}[title={Example Prompt for Experience Rewriting},breakable,]
  You are an expert AI assistant tasked with rewriting and reorganizing context content to make it more relevant and actionable for the current task.\\
  
  Your task is to take the original context (containing multiple experiences) and rewrite it as a cohesive, task-specific guidance that directly addresses the current situation.\\
  
  REWRITING GUIDELINES:\\
  $\bullet$ RELEVANCE FOCUS: Emphasize the most relevant aspects of each experience. Prioritize the most relevant experiences. Use clear, direct language.\\
  $\bullet$ ACTIONABLE INSIGHTS: Extract specific, actionable guidance. Make the context immediately actionable\\
  $\bullet$ COHERENT NARRATIVE: Create a flowing narrative rather than disconnected tips\\
  $\bullet$ SITUATIONAL AWARENESS: Adapt the guidance to the current situation\\
  
  \# Current Task/Query\\
  {current\_query}\\
  
  \# Current Trajectory\\
  {current\_context}
  
  \# Original Context Content (Multiple Experiences)\\
  {original\_context}\\
  
  OUTPUT FORMAT:\\
  Provide the rewritten context:\\
  \verb|```|json\\
  $\left\{ \left\{ \right.\right.\\
    \left. \right. \; ``\text{rewritten\_context}": ``\text{A cohesive, task-specific context message that reorganizes and adapts the} \\
    \left. \right. \; \text{original experiences for the current task. This should be written as a unified guidance rather than} \\
    \left. \right. \; \text{separate experience items.}",\\
  \left. \left. \right\} \right\}$\\
  \verb|```|\\
  
  Guidelines:\\
  - Rewrite as a unified, flowing guidance\\
  - Adapt terminology and examples to match the current task domain\\
  - Consolidate overlapping insights into coherent recommendations\\
  - Prioritize experiences most relevant to the current situation\\
  - Make the guidance feel custom-written for this specific task\\
\end{tcolorbox}
\end{tabular}
\caption{Example prompt for experience rewrting.}
\label{tab:rewrite_prompt}
\end{table*}

\end{document}